\documentclass[journal]{IEEEtran}
\usepackage{graphicx}
\usepackage{graphics}
\usepackage{amssymb}
\usepackage{color}
\usepackage{ulem}
\usepackage{array}
\usepackage{multirow}
\usepackage{amsmath}
\usepackage{bm}
\usepackage{cite}
\usepackage{float}
\usepackage[breaklinks=true,bookmarks=false, hidelinks]{hyperref}

\newlength\savedwidth
\newcommand\whline{\noalign{\global\savedwidth\arrayrulewidth
                            \global\arrayrulewidth 1pt}%
                   \hline
                   \noalign{\global\arrayrulewidth\savedwidth}}

\ifCLASSINFOpdf
\else
\fi
\begin{document}
%
\title{Deep Multi-Branch Aggregation Network for Real-Time Semantic Segmentation in Street Scenes}
%
%
%

\author{Xi~Weng,
        Yan~Yan, ~\IEEEmembership{Member,~IEEE,}
        Genshun~Dong,
       {Chang~Shu,}
        Biao~Wang, \\
        Hanzi~Wang,~\IEEEmembership{Senior Member,~IEEE},
        and~Ji~Zhang,~\IEEEmembership{Senior Member,~IEEE}
\thanks{This work was partly supported by the National
Natural Science Foundation of China under Grants 62071404,
U21A20514, and 61872307, by the Open Research Projects of Zhejiang
Lab under Grant 2021KG0AB02, by the Natural Science Foundation of
Fujian Province under Grant 2020J01001, and by the Youth Innovation
Foundation of Xiamen City under Grant 3502Z20206046. \emph{(Corresponding authors: Yan Yan; Biao Wang.)}}
\thanks{X.~Weng, Y.~Yan, G.~Dong, and H.~Wang are with the Fujian Key Laboratory of Sensing and
Computing for Smart City, School of Informatics, Xiamen University, China (e-mail: xweng@stu.xmu.edu.cn; yanyan@xmu.edu.cn; gshdong@qq.com;
hanzi.wang@xmu.edu.cn).}
\thanks{C.~Shu is with the School of Information and Communication Engineering, University of Electronic Science and Technology of China, China (e-mail: changshu@uestc.edu.cn).}
 \thanks{B.~Wang is with Zhejiang Lab, China (e-mail: wangbiao@zhejiang\-lab.com).}
  \thanks{J.~Zhang is with University of Southern Queensland, Australia \&
Zhejiang Lab, China (e-mail: zhangji77@gmail.com).}
 }

%
%

\markboth{Journal of \LaTeX\ Class Files}%
{Shell \MakeLowercase{\textit{et al.}}: Bare Demo of IEEEtran.cls for IEEE Journals}
%



\maketitle

\begin{abstract}
Real-time semantic segmentation, which aims to achieve high segmentation accuracy at real-time inference speed, has received substantial attention over the past few years.
However, many state-of-the-art real-time semantic segmentation methods tend to sacrifice some spatial details or contextual information for fast inference, thus leading to degradation in segmentation quality.
In this paper, we propose a novel Deep Multi-branch Aggregation Network (called DMA-Net) based on the encoder-decoder structure to perform real-time semantic segmentation in street scenes. Specifically, we first adopt ResNet-18 as the encoder to efficiently generate various levels of feature maps from different stages of convolutions. Then, we develop a Multi-branch Aggregation Network (MAN) as the decoder to effectively aggregate different levels of feature maps and capture the multi-scale information. {In MAN, a lattice enhanced residual block is designed to enhance feature representations of the network by taking advantage of the lattice structure. Meanwhile, a feature transformation block is introduced to explicitly transform the feature map from the neighboring branch before feature aggregation. Moreover, a global context block is used to exploit the global contextual information.} These key components are tightly combined and jointly optimized in a unified network.
Extensive experimental results on the challenging Cityscapes and CamVid datasets demonstrate that our proposed DMA-Net respectively obtains 77.0\% and 73.6\% mean Intersection over Union (mIoU) at the inference speed of 46.7 FPS and 119.8 FPS by only using a single NVIDIA GTX 1080Ti GPU.
This shows that DMA-Net provides a good tradeoff between segmentation quality and speed for semantic segmentation in street scenes.
\end{abstract}


%
\IEEEpeerreviewmaketitle

\section{Introduction}
%
%
%
%
\IEEEPARstart{S}{emantic} segmentation, which predicts the semantic label of each pixel in an image, is a fundamental and challenging task in street scene understanding. During the past few decades, semantic segmentation in street scenes has attracted increasing attention, mainly due to its important role in autonomous driving systems \cite{ITS1,SwiftNet,ITS2,ITS3}. Generally, these systems demand fast inference speed for interaction and response.

\begin{figure}[!t]
\begin{center}
  \includegraphics[width=1.0\linewidth]{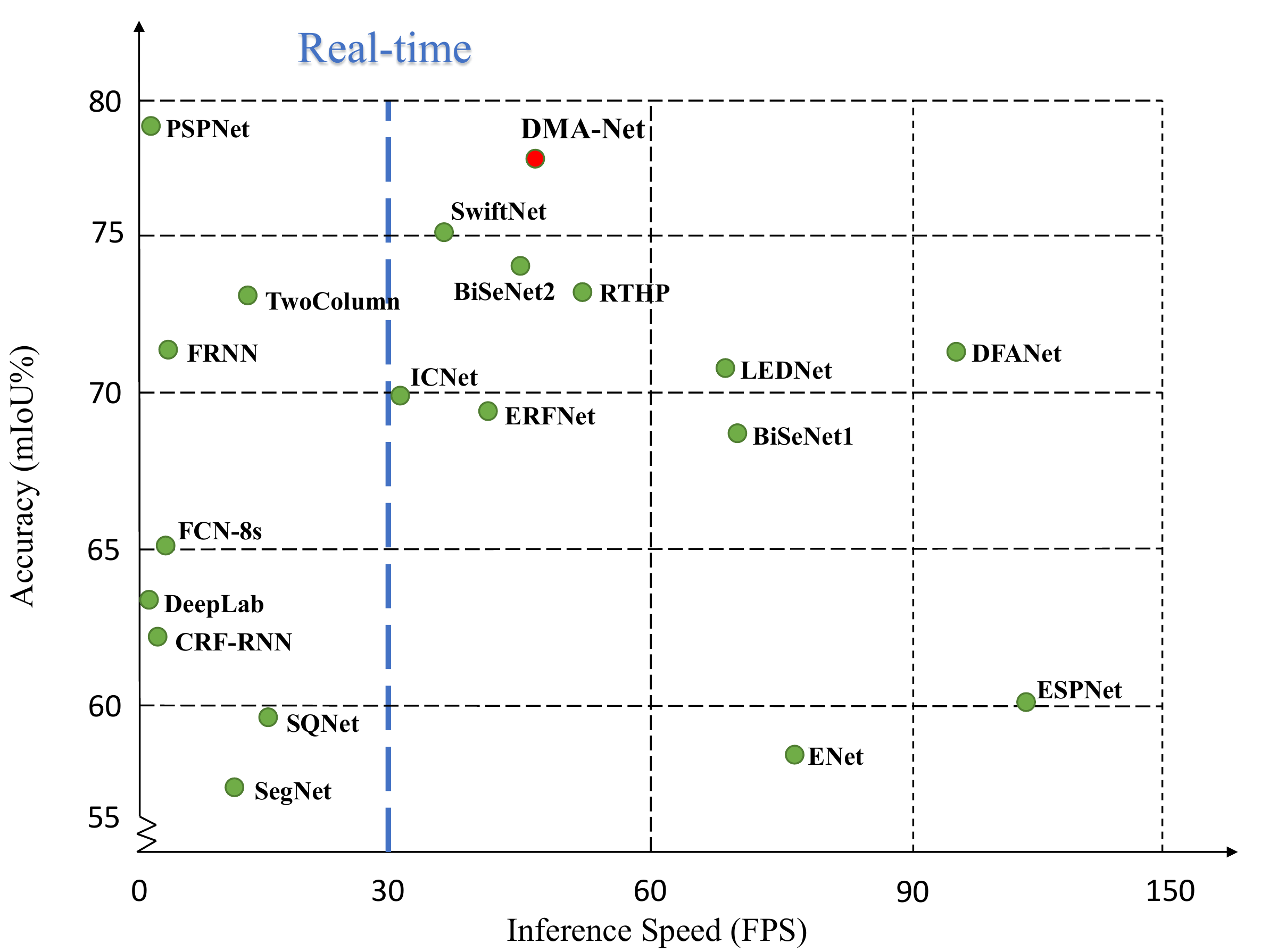}
\end{center}
  \caption{Accuracy (mIoU) and inference speed (FPS) obtained by several state-of-the-art semantic segmentation methods, including SwiftNet\cite{SwiftNet}, PSPNet\cite{PSPNet}, ENet \cite{ENet}, ERFNet \cite{erfnet}, BiSeNet \cite{BiSeNet}, ICNet \cite{ICNet}, LEDNet \cite{LEDNet},  RTHP \cite{pre-work},  DFANet \cite{DFANet}, ESPNet \cite{espnet}, FCN-8s \cite{FCN}, DeepLab \cite{deeplab}, CRF-RNN \cite{CRF-RNN}, SegNet \cite{SegNet}, SQNet \cite{SQNet}, FRRN \cite{FRRN}, TwoColumn \cite{TwoColumn}, and the proposed DMA-Net on the  Cityscapes test set.}
\label{first-fig}
\end{figure}

{Street scene images are often captured by a surveillance camera mounted behind the windshield of a driving car. Generally, images in street scene datasets (such as Cityscapes \cite{cityscapes} and CamVid \cite{camvid}) contain different kinds of objects (e.g., road, car, and building). Compared with the objects in natural scenes, some objects in street scenes are visually similar {(such as building vs. wall, and truck vs. bus)}. How to distinguish similar objects is of great importance for street scene understanding and plays a critical role in achieving good segmentation accuracy.}

{{Benefiting from the outstanding performance of Deep Convolutional Neural Network (DCNN), a large number of semantic segmentation methods \cite{PSPNet,RefineNet,deeplabv2,deeplabv3+} have been proposed} and shown significant performance improvements in terms of segmentation accuracy, {especially for distinguishing similar objects in street scenes}.
The success of the above methods relies largely on sophisticated DCNN models (such as Xception \cite{xception} and ResNet-101 \cite{resnet}) as the {backbone networks} to capture low-level spatial details and high-level semantics. Unfortunately, these DCNN models usually involve heavy computational operations and high memory consumption.} As a consequence, although remarkable progress has been made by these methods, their high computational costs and memory requirements inhibit the deployment of semantic segmentation in many real-world applications with limited power resources (such as self-driving cars and driver assistance systems).

{To achieve fast inference speed, a variety of real-time semantic segmentation methods \cite{ENet,espnet,BiSeNet,ICNet,LEDNet,pre-work,DFANet,erfnet} have been developed
by leveraging lightweight networks (such as MobileNetV2 \cite{mobilenetv2} and ShuffleNet \cite{shufflenet}) as the backbone networks.
However, the feature extraction capability of lightweight networks is often inferior, and thus these networks are difficult to extract
feature maps with rich spatial and contextual information for
pixel-level classification. Therefore, the accuracy
of these methods in segmenting similar objects in street scenes is greatly affected.} {Fig. \ref{first-fig} shows the accuracy {(mean Intersection over Union (mIoU))} and inference speed {(Frames Per Second (FPS))} obtained by several state-of-the-art semantic segmentation methods on the Cityscapes test set. } Obviously, different from the rapid development of high-quality semantic segmentation methods, research towards real-time semantic segmentation {in street scenes} without reducing too much accuracy is still left behind.

Recently, some methods, such as Bilateral Segmentation Network (BiSeNet) \cite{BiSeNet} and Deep Feature
Aggregation Network (DFANet) \cite{DFANet}, have been developed in pursuit of high segmentation accuracy at real-time inference speed.  {BiSeNet employs a two-branch DCNN model to combine the spatial and semantic information.} Nevertheless, the lack of communication between two branches may weaken the learning capacity of the model.
DFANet makes use of deep feature aggregation to address real-time semantic segmentation on high-resolution images, where the {feature maps} are concatenated at both the network-level and the stage-level.
{However, simple aggregation operations (such as the element-wise addition and the channel-wise concatenation) \cite{pre-work,DFANet} are not optimal since the feature maps from the encoder  have a gap. These operations may cause feature interference, and thus the decoder cannot faithfully pay attention to objects at different scales, leading to a performance decrease. This issue is more pronounced in street scenes, which usually cover different scales of objects.}

In the light of the above issues, we propose a novel Deep Multi-branch Aggregation Network, called DMA-Net, based on the encoder-decoder structure
for real-time semantic segmentation in street scenes.
Specifically, we adopt a lightweight network (i.e., ResNet-18 \cite{resnet}) as the encoder and develop a Multi-branch Aggregation
Network (MAN) as the decoder.
{In MAN, a Lattice Enhanced Residual Block (LERB) consisting of two lattice structures is designed to combine the spatial and contextual enhanced blocks in each branch of MAN. In particular, we leverage two {weight learning blocks} to adjust the weights of two lattice structures adaptively.} {Meanwhile, a Feature Transformation Block (FTB), which emphasizes the important information while ignoring  the irrelevant information in the feature maps, is introduced to explicitly transform the feature map from the
neighboring branch before feature aggregation.}
Moreover, a Global Context Block (GCB) is employed to capture the rich global contextual information, which is critical for semantic segmentation.

In summary, our main contributions of this paper are summarized as follows:

\begin{itemize}
\item {We develop LERB to effectively enhance both
spatial details and contextual information of feature maps from the encoder.
   In particular, the lattice structures in LERB {allow} the potential of various combinations of enhanced blocks, greatly enlarging the representation space of LERB in an
 efficient manner.
 {Therefore, the problem of inferior feature extraction capability of the lightweight backbone network is significantly solved, improving the performance of segmenting similar objects.}}


\item {We propose FTB to generate the transformed feature maps based on a transformation tensor at a relatively small computational cost. In this way, the gap between different levels of feature maps is largely mitigated. {As a result, the problem of feature interference between high-level and low-level feature maps is alleviated,} and these feature maps can be appropriately aggregated.}

\item {The key components (i.e., ResNet-18, LERB, FTB, and GCB) are tightly combined and jointly optimized in DMA-Net to achieve real-time semantic segmentation in street scenes. Our proposed DMA-Net obtains 77.0\% and 73.6\% mIoU on the challenging Cityscapes and CamVid test datasets at the speed of 46.7 FPS and 119.8 FPS, respectively (only a single NVIDIA GTX 1080Ti GPU is used). These results demonstrate that our proposed {DMA-Net} is able to make a good tradeoff between accuracy and speed for semantic segmentation in street scenes.}
\end{itemize}

The rest of this paper is organized as follows. First,
we review the related work in Section \ref{sec:related-work}. Then, we
 describe the proposed {DMA-Net} in detail in Section \ref{sec:approach}. Next, we give ablation studies and show experimental results on two challenging street scene semantic segmentation datasets in Section \ref{sec:experiments}. Finally, we draw our conclusion in Section \ref{sec:conclusion}.


\section{Related Work}
\label{sec:related-work}

DCNN has made great success in various computer vision tasks, since its outstanding achievement on the large-scale image classification task \cite{alexnet}. In recent years, a series of DCNN-based semantic segmentation methods have been developed and achieved excellent performance on the benchmark datasets. In this section, we briefly review some state-of-the-art semantic segmentation methods, including {high-quality} methods and real-time ones.

\subsection{{High-Quality} Semantic Segmentation Methods}
Fully Convolutional Network (FCN) \cite{FCN} is the pioneering semantic segmentation method. FCN replaces the fully-connected layers of the classification networks with the convolutional layers, and it forms the foundation of modern semantic segmentation methods.  To generate dense feature maps, FCN makes use of skip connections to combine the coarse and fine {feature maps}.
Ronneberger \textit{et al}. \cite{U-Net}  propose a U-shape Network (U-Net), which consists of an encoder and a decoder. The encoder gradually increases the receptive {fields} to capture the {contextual} information, while the decoder recovers the spatial information from the outputs of {the} encoder in a layer-by-layer manner. 
DeepLab \cite{deeplab} introduces the atrous convolution \cite{atrous-convolution} to enlarge the receptive fields of the network without increasing the number of parameters.
DeepLabv3+ \cite{deeplabv3+} also adopts the encoder-decoder structure, where a network similar to
DeepLabv3 \cite{deeplabv3} is employed to encode the contextual information while a simple decoder is leveraged to refine the segmentation accuracy (especially near the object boundaries). {Multi-path Refinement Network (RefineNet) \cite{RefineNet} refines high-level {feature maps} by using fine-grained low-level {feature maps} based on a generic multi-path framework.}

The existence of objects at multiple scales in street scenes raises a great challenge in semantic segmentation. To address this challenge, {a standard way is to perform segmentation on multiple re-scaled versions of the same input {image} and then aggregate the output feature maps. Although such a way can boost the segmentation accuracy, {it usually significantly increases the computational burden \cite{re-scaled-versions}.}} DeepLabv2 \cite{deeplabv2} {develops an} Atrous Spatial Pyramid Pooling (ASPP) module to
robustly segment  {multi-scale objects}.
ASPP extracts {feature maps} in  {multiple parallel} atrous convolution branches with different sampling rates, thus capturing objects
and  {contextual information} at different scales. Similarly,  {Pyramid Scene Parsing Network (PSPNet)} \cite{PSPNet} aggregates the contextual information from different regions based on a pyramid network structure.
Context Encoding Network (EncNet) \cite{EncNet} exploits the global contextual information through a context encoding module to enlarge the receptive fields and segment multi-scale objects.
To refine the outputs,  {some} methods \cite{deeplab}\cite{CRF-RNN}  {also} employ the
probabilistic graphical model, such as Conditional Random Fields (CRF) \cite{CRF}, as a post-processing step to improve the segmentation accuracy of object boundaries.

Recently, the self-attention mechanism has been adopted in several state-of-the-art methods. Dual Attention Network (DANet) \cite{DANet} develops a position and channel attention module to improve the segmentation accuracy by adaptively capturing and aggregating the contextual information. Expectation-Maximization Attention Network (EMANet) \cite{EMANet} computes a compact basis set to reduce the computational complexity of semantic segmentation by using an expectation-maximization iteration manner.

The aforementioned methods show high segmentation {accuracy} on various benchmark datasets. Many methods (such as RefineNet \cite{RefineNet} and U-Net \cite{U-Net}) adopt the encoder-decoder structure.
Unfortunately, they generally suffer from heavy computational costs and long inference time, due to the large number of network parameters or the large-scale {floating-point} operations, or both. In this paper, DMA-Net is also based on the encoder-decoder structure. However, compared with symmetric encoder-decoder structures used in U-Net and RefineNet, {DMA-Net is much more lightweight and specifically designed for real-time semantic segmentation in street scenes.}

\subsection{Real-Time Semantic Segmentation Methods}

Real-time semantic segmentation methods aim to generate high-quality prediction at fast inference speed (e.g., more than {30} FPS). Segmentation Neural Network (SegNet) \cite{SegNet} is the early real-time semantic segmentation method, which removes the fully-connected layers in the network to obtain a small architecture and utilizes the max pooling indices to upsample the {feature maps}. Efficient Neural Network (ENet) \cite{ENet} designs a compact encoder-decoder structure, where early downsampling is employed to make it suitable for the low-latency semantic segmentation task. However, ENet cannot robustly segment large objects due to the relatively small receptive  {fields} used in the compact architecture.
Efficient Spatial Pyramid Network (ESPNet) \cite{espnet} proposes an efficient spatial pyramid module, where the standard convolution is decomposed into point-wise convolutions and a spatial pyramid of dilated convolutions. Hence, the computational complexity of the model is reduced. Similarly, Efficient Residual Factorized Network (ERFNet) \cite{erfnet}  {designs} a novel convolutional layer, which utilizes residual connections and factorized convolutions to efficiently perform semantic segmentation.  The above methods usually compromise spatial details or contextual information to achieve fast inference speed. Such a manner leads to poor segmentation results. Therefore, compared with high-quality semantic segmentation methods, the segmentation accuracy of these methods is still far from being satisfactory.

Recently, the multi-branch framework has drawn much interest. For example,
Zhao \textit{et al.} \cite{ICNet} propose an Image Cascade Network (ICNet) based on {the} simplified version of  {PSPNet} and cascade networks. ICNet combines the semantic information from low-resolution images and the detailed spatial information from high-resolution images. BiSeNet \cite{BiSeNet} adopts a two-branch DCNN structure to respectively  {encode} the spatial  and semantic information, so as to improve both the inference speed and segmentation accuracy. Note that BiSeNet explores the spatial details and the semantic information separately. The lack of communication between branches may influence the learning ability of the DCNN model.
To address the above problem, DFANet \cite{DFANet} employs a feature reuse strategy to make a balanced tradeoff between accuracy and speed. However, DFANet aggregates {feature maps} at the different levels by a simple network structure, thereby ignoring the differences between them.

{In this paper, the proposed {DMA-Net} also takes advantage of the multi-branch framework.
However, different from the above methods, DMA-Net progressively aggregates the feature maps from the high-level branch to the low-level branch based on an elaborately-designed lightweight decoder MAN (mainly consisting of LERB, FTB, and GCB).
Therefore, DMA-Net is able to effectively and efficiently segment objects in complex street scenes.
 Moreover, {DMA-Net}
makes use of different levels of feature maps from different stages of ResNet-18 as the inputs for multiple branches in MAN, where a principal loss or an auxiliary loss is specifically employed to supervise the output of each branch. As a result, each branch focuses on capturing the semantic information at a certain scale.}

\begin{figure*}
\begin{center}
\includegraphics[width=1.0\linewidth]{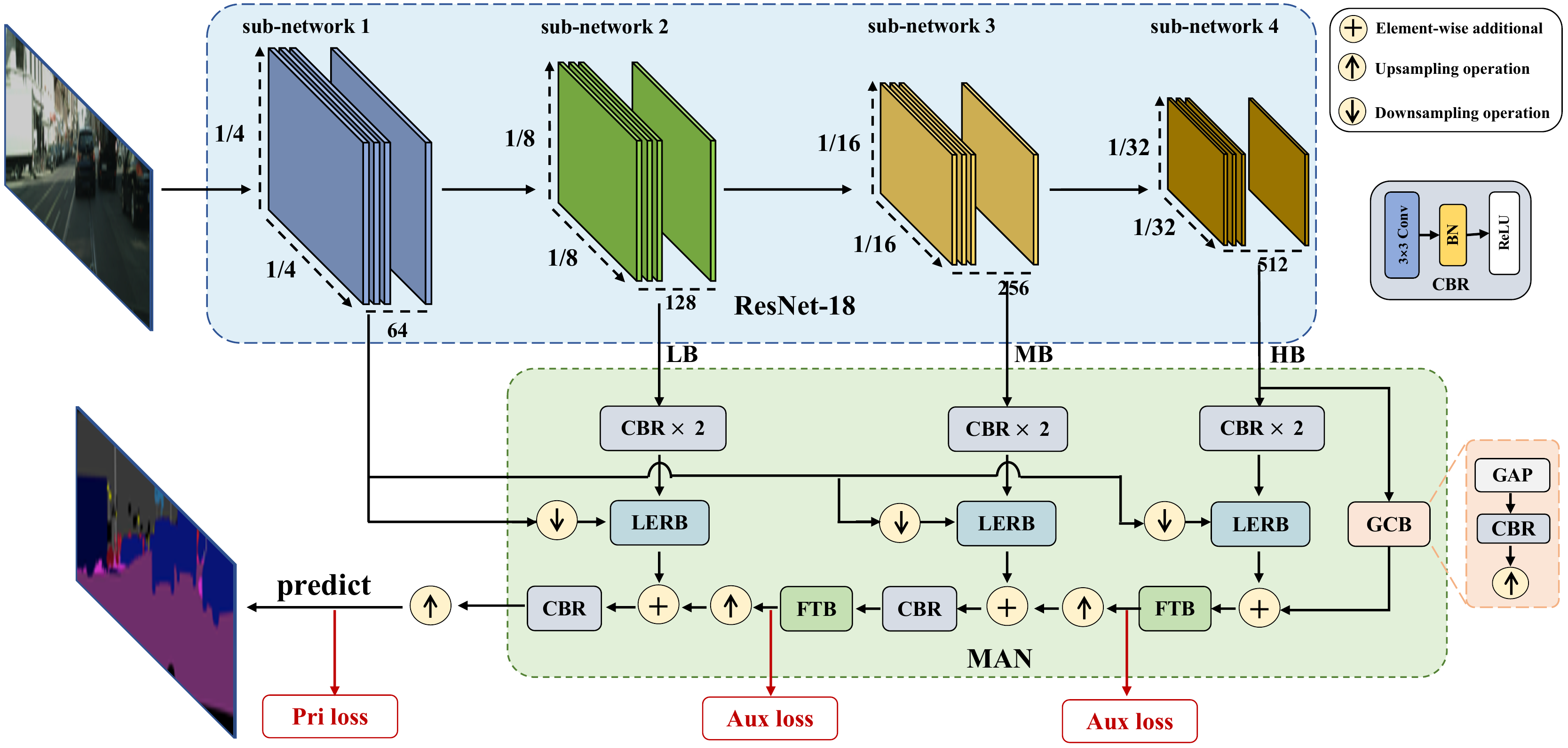}
\end{center}
  \caption{{The overall framework of the proposed DMA-Net. ResNet-18 extracts different levels of feature maps. MAN aggregates the feature maps from ResNet-18 to generate the final prediction. In the figure, {CBR} denotes the Conv-BN-ReLU module. ``GAP'' denotes the global average pooling operation. ``Pri loss'' and ``Aux loss'' represent the principal loss and the auxiliary loss, respectively.}}
\label{fig:architecture-network}
\end{figure*}

\section{The Proposed Method}
\label{sec:approach}
In this section, we present the proposed {DMA-Net} in detail.  We first give an overview of DMA-Net in Section \ref{approach-overview}. Then, we introduce each component of {DMA-Net} from Section \ref{approach-ResNet-18} to Section \ref{approach-MBA}. {Next, we give the joint loss in Section \ref{loss}. Finally, we present some discussions about our DMA-Net in Section \ref{discuss}.}

\subsection{Overview}
\label{approach-overview}

{DMA-Net consists of two main parts: ResNet-18 and a Multi-branch Aggregation Network (MAN). In particular, a Lattice Enhanced Residual Block (LERB), a Feature Transformation Block (FTB), and a Global Context Block (GCB) are developed in MAN.}


{The overall framework of  {DMA-Net} is illustrated in Fig. \ref{fig:architecture-network}. An image \textbf{I} $\in  \mathbb{R}^{H \times W \times C } $ is taken as the input of ResNet-18, where $H$, $W$, and $C$ represent the {height}, the width, and the number of channels of the image \textbf{I}, respectively. First, ResNet-18 efficiently downsamples the input image by several consecutive convolutional blocks to generate different {levels of} feature maps.
Then, as the core of  DMA-Net, MAN takes the feature maps from different stages of ResNet-18 as the inputs, and progressively performs {feature aggregation} from the high-level branch to the low-level branch.} {In MAN, LERB effectively enhances feature representations of the network, while FTB greatly reduces the semantic gap between feature maps before feature aggregation. In addition, instead of relying on multi-scale input images or a specifically-designed multi-scale module, MAN not only exploits the multi-scale information by recursively aggregating feature maps from the high-level branch to the low-level branch, but also explicitly adopts both the principal loss and the auxiliary losses.}


DMA-Net is a lightweight encoder-decoder network. On the one hand, we employ ResNet-18, which is much simpler than complex DCNN models (such as ResNet-101 and Xception) used in high-quality semantic segmentation methods, as the encoder to ensure high inference speed. On the other hand, we develop MAN as the decoder with a small amount of network parameters to efficiently and effectively combine spatial details and contextual information. Therefore, DMA-Net can achieve a good balance between accuracy and inference speed.


\subsection{{ResNet-18}}
\label{approach-ResNet-18}


{An encoder plays a critical role in basic feature extraction of the input images.
In this paper, we adopt ResNet-18 (pre-trained with ImageNet \cite{ImageNet}) as our encoder. The input images are downsampled by using a max-pooling layer at the earlier layer of ResNet-18. Moreover,  ResNet-18 is composed of a small number of layers in the network. Therefore, ResNet-18 has the distinct advantage of high efficiency with fast speed and small memory consumption.} 

{Specifically, we remove all the network layers (including the pooling layers and the fully-connected layers, etc.) after the last residual building block of ResNet-18 to obtain a simplified version of ResNet-18. Hence, the network architecture of the simplified version of ResNet-18 consists of a standard $7 \times 7$ convolutional layer, a $3 \times 3$ max-pooling layer, and eight $3 \times 3$ residual  {building} blocks. The eight residual  {building} blocks can be divided into {four} sub-networks (i.e., sub-network 1 to sub-network 4), according to the size of the output feature maps, as shown in Fig.~\ref{fig:architecture-network}. Generally, the size of the output feature maps is reduced to one half after passing through  {each} sub-network.
Therefore, we can obtain four different levels of feature maps (whose sizes correspond to $1/4$, $1/8$, $1/16$, and $1/32$ of the original image size)  from four sub-networks.}

\begin{figure}[!t]
\begin{center}
  \includegraphics[width=0.7\linewidth]{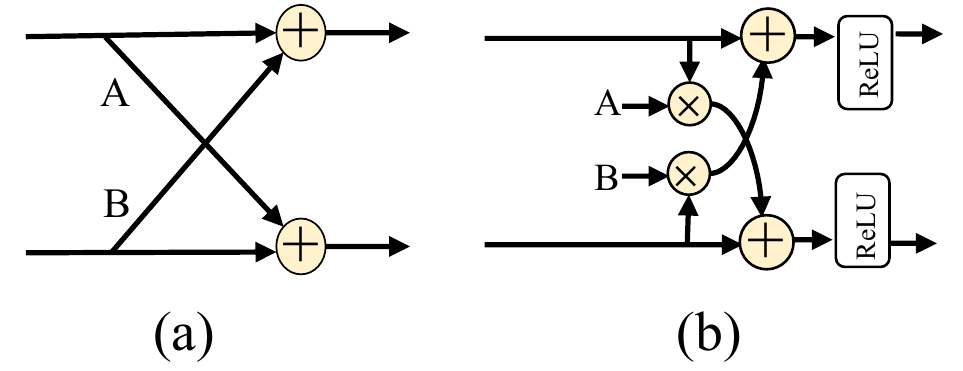}
\end{center}
   \caption{{The network architecture of (a) a standard lattice filter with 2-channel filters, and (b) our lattice structure.}}
\label{fig:LS}
\end{figure}

\begin{figure*}[!t]
\begin{center}
  \includegraphics[width=1.0\linewidth]{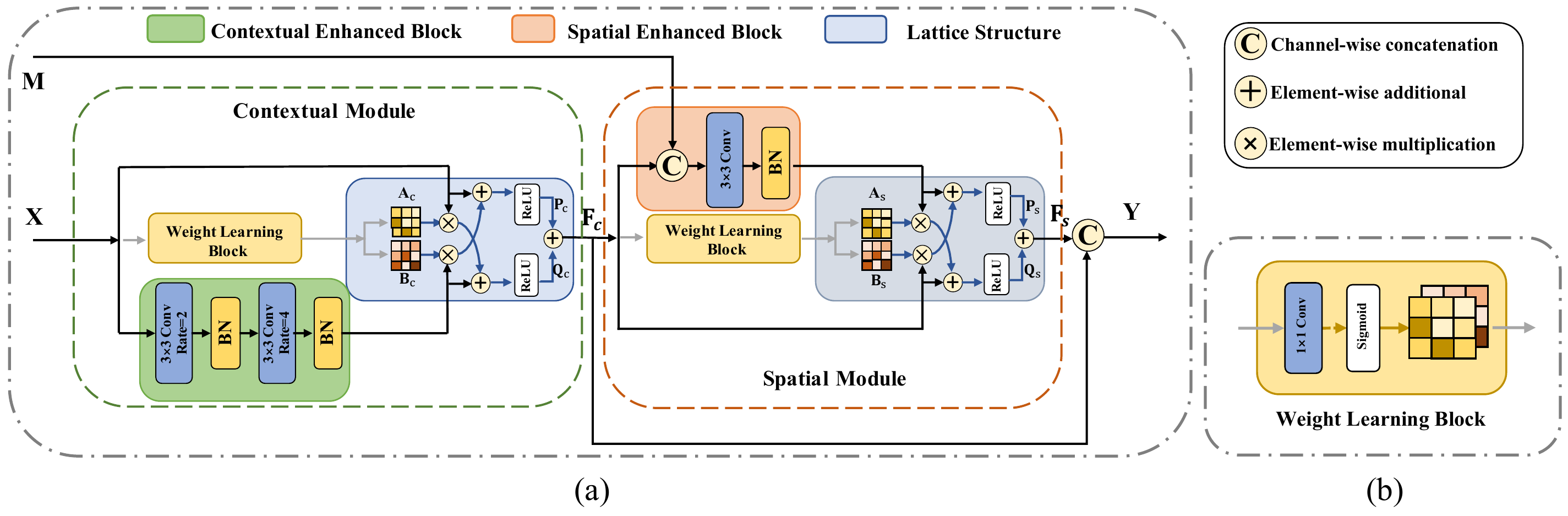}
\end{center}
   \caption{{The network architecture of (a) LERB  and  (b) Weight Learning Block. In the figure, ``Rate" means the atrous rate. ``BN" denotes the batch normalization layer. ``ReLU'' and ``Sigmoid'' indicate the ReLU and Sigmoid activation functions, respectively.}}
\label{fig:ERB}
\end{figure*}

\subsection{{Multi-branch Aggregation Network (MAN)}}
\label{approach-MBA}
{Compared with complex DCNN models, the feature extraction capability of ResNet-18 is inferior. To achieve a good balance between segmentation accuracy and inference speed, we resort to an elaborately-designed decoder MAN to aggregate different levels of feature maps for semantic segmentation in street scenes. Thus, the spatial and contextual information can be effectively combined in the decoder.
In particular, multiple {Conv-BN-ReLU} modules (each {Conv-BN-ReLU} module includes a $3 \times 3$ convolutional layer followed by a Batch Normalization (BN) layer and a ReLU activation function) are used in MAN to reduce the number of feature channels. Such a way ensures the small computational cost of MAN.}

The network architecture of MAN
is shown in Fig.~\ref{fig:architecture-network}. {We can see that, the four different levels of feature maps from four sub-networks of ResNet-18 are used as the inputs of MAN.} {To be specific, the feature maps, whose sizes are $1/8$, $1/16$, and $1/32$ of the original input image size, are respectively fed into three branches, including a Low-level Branch (LB), a Mid-level Branch (MB), and a High-level Branch (HB). For each branch, we first employ two {Conv-BN-ReLU} modules
to reduce the dimension of  the feature map.
Then, LERB is designed to improve the {feature representations}, given the feature maps from {the Conv-BN-ReLU module} and the first sub-network of ResNet-18.}
Finally, the output feature map of LERB is combined with the transformed feature map based on FTB from the neighboring branch.  {Note that, in HB, the {last} downsampled feature maps obtained from ResNet-18 are also fed into GCB to model the {global contextual dependency}, which can {provide the rich} high-level
contextual information for MAN. The outputs of these branches are progressively aggregated to obtain the final predicted results.} 

It is worth pointing out that our proposed MAN is able to effectively and efficiently capture the multi-scale information. { As it is well known \cite{re-scaled-versions, deeplabv3, pre-work}}, one problem in the application of DCNN to semantic segmentation is the difficulty of using a single scale to perform pixel-level dense prediction, because of the existence of objects at multiple scales.
Hence, how to accurately capture the multi-scale object information while maintaining fast inference speed of the network is a great challenge. Traditional methods either rely on multiple re-scaled versions of the input images \cite{re-scaled-versions} or use an additional multi-scale module (such as ASPP \cite{deeplabv3} or DASPP \cite{pre-work}) to tackle the multi-scale problem. However, {such manners \cite{deeplabv3, pre-work} usually} bring additional consumption in terms of both computational complexity and memory requirement.

Different from the above methods, our proposed MAN recursively aggregates the multi-level information from the different branches to obtain the segmentation results. In MAN, each branch tackles the {feature map} at a certain size from the sub-networks of ResNet-18 and is trained by using a principal loss or an auxiliary loss. This enables MAN to successfully deal with the multi-scale problem of semantic segmentation.


{In the following, we respectively introduce {three key components of MAN (i.e., LERB, FTB, and GCB)} in detail.}

\subsubsection{{Lattice Enhanced Residual Block (LERB)}}
\label{approach-MRB}
{In this paper, inspired by the residual building blocks \cite{resnet} and the lattice filter \cite{lattice}, we develop LERB to enhance feature representations in each branch. The structure of the lattice filter, also called as X-section, is the physical  topology of an all-pass filter with the butterfly structure, which decomposes the input signal to multi-order representations \cite{lattice}. Fig.~\ref{fig:LS} shows the network architecture of a standard lattice filter and the lattice structure used in our method.}

The network architecture of LERB is shown in Fig.~\ref{fig:ERB}. LERB mainly consists of a contextual module and a spatial module to enhance the contextual information and spatial details, respectively.  Specifically, the input feature map {$\textbf{X}\in\mathbb{R}^{H^{l}\times W^{l}\times C^{l}}$} is fed into the {contextual module consisting of a {contextual enhanced block, a weight learning block, and a lattice structure. The contextual enhanced block} contains two $3 \times 3$ convolutional layers followed by a BN layer. Here, the atrous rates of two convolutional layers are respectively set to 2 and 4 to capture sufficient contextual information.
Meanwhile, {the weight learning block} (consisting of a $1 \times 1$ convolutional layer and a Sigmoid activation function) is adopted to adaptively learn two weight tensors (i.e., $\textbf{A}_c\in\mathbb{R}^{H^{l}\times W^{l}\times 1}$ and $\textbf{B}_c\in\mathbb{R}^{H^{l}\times W^{l}\times 1}$), which are used for the {lattice structure.}  The nonlinear function induced by {the contextual enhanced block} is denoted as $\mathrm{C}(\cdot)$. Therefore, the two output feature maps in the {lattice structure} are formulated as
{
\begin{equation}
\begin{split}
\textbf{P}_c&= \sigma(\textbf{X} +  \eta(\textbf{B}_c)\otimes\mathrm{C}(\textbf{X})), \\
\textbf{Q}_c&= \sigma(\eta(\textbf{A}_c)\otimes\textbf{X} + \mathrm{C}(\textbf{X})),
\end{split}
\end{equation}}
where $\textbf{P}_c\in\mathbb{R}^{H^{l}\times W^{l}\times C^{l}}$ and $\textbf{Q}_c\in\mathbb{R}^{H^{l}\times W^{l}\times C^{l}}$} represent the intermediate feature maps. $\sigma(\cdot)$ denotes the ReLU activation function. `$\otimes$' means the element-wise multiplication operation. {$\eta(\cdot)$ indicates the broadcast operation, where the weights are broadcast (copied) along the channel dimension.}

{The output feature map {$\textbf{F}_c\in\mathbb{R}^{H^{l}\times W^{l}\times C^{l}}$} from {the lattice structure} can be obtained as
\begin{equation}
\textbf{F}_c= \textbf{P}_c \oplus \textbf{Q}_c,
\end{equation}
where `$\oplus$' represents the element-wise addition operation.}


\begin{figure}[t]
\begin{center}
  \includegraphics[width=0.7\linewidth]{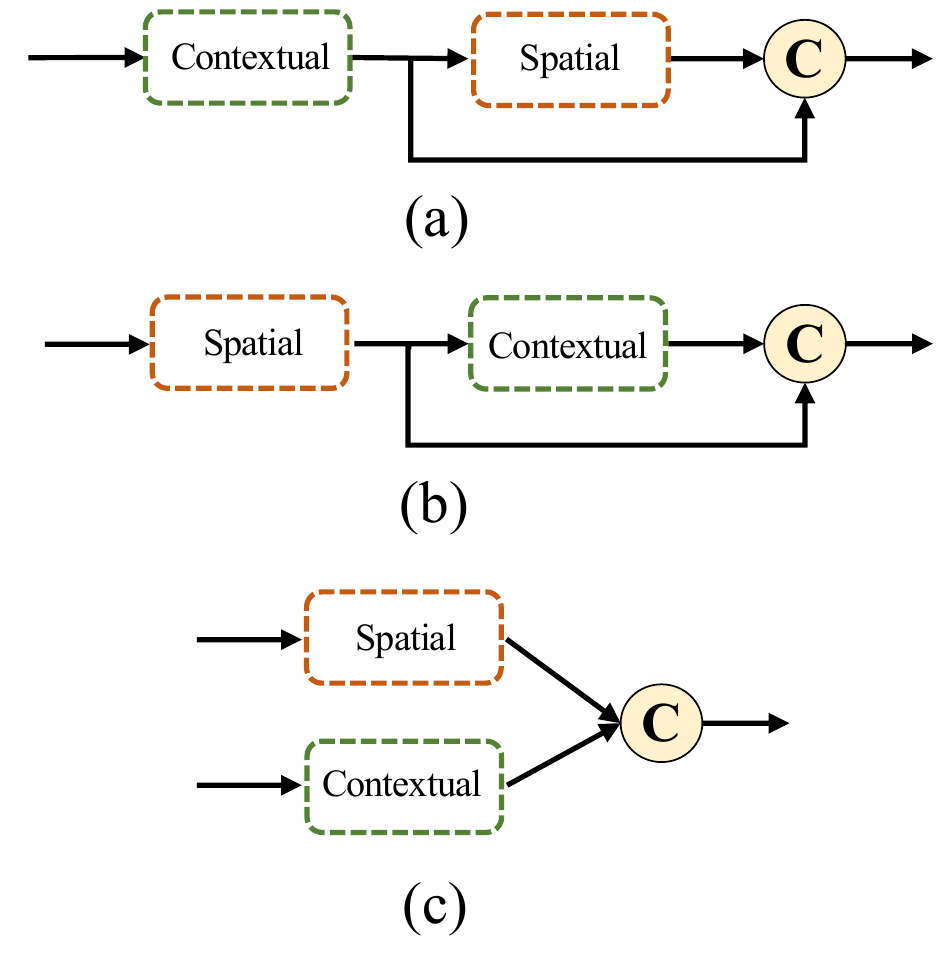}
\end{center}
   \caption{{Various ways of combining the contextual module and the spatial module. (a) the spatial module after the contextual module in serial (i.e., LERB), (b) the contextual module after the spatial module in serial, and (c) the spatial module and the contextual module in parallel. In the figure, ``Contextual" and ``Spatial" respectively represent the contextual module and the spatial module, respectively.}}
\label{fig:LERB-three-different}
\end{figure}

Then, $\textbf{F}_c$ is passed through a {spatial module consisting of a {spatial enhanced block, a weight learning block, and a lattice structure}. Meanwhile, the downsampled feature map $\textbf{M}$ from the first sub-network of ResNet-18, which has the same size as $\textbf{F}_c$, is also used as the input of {the spatial enhanced block}. In {the spatial enhanced block}, $\textbf{F}_c$ and $\textbf{M}$ are first concatenated along the channel dimension. By concatenating the feature maps from the first lattice structure and the downsampled ones from the first sub-network of ResNet-18, we are able to enhance spatial details in the {spatial module}. Based on the concatenated feature map, a $3 \times 3$ convolutional layer followed by a BN layer is used to enhance feature spatial representations. The nonlinear function induced
by {the spatial enhanced block} is denoted as $\mathrm{S}(\cdot)$. {Meanwhile}, $\textbf{F}_c$ is also used to learn two weight tensors (i.e., $\textbf{A}_s\in\mathbb{R}^{H^{l}\times W^{l}\times 1}$ and $\textbf{B}_s\in\mathbb{R}^{H^{l}\times W^{l}\times 1}$) by a {weight learning block}. Therefore, the two output feature maps in the {lattice structure} are formulated as
{
\begin{equation}
\begin{split}
\textbf{P}_s&= \sigma(\eta(\textbf{B}_s)\otimes\textbf{F}_c +  \mathrm{S}(\mathrm{concat}(\textbf{F}_c, \textbf{M}))), \\
\textbf{Q}_s&= \sigma(\textbf{F}_c + \eta(\textbf{A}_s)\otimes\mathrm{S}(\mathrm{concat}(\textbf{F}_c, \textbf{M}))),
\textbf{}
\end{split}
\end{equation}}
{where $\textbf{P}_s\in\mathbb{R}^{H^{l}\times W^{l}\times C^{l}}$ and $\textbf{Q}_s\in\mathbb{R}^{H^{l}\times W^{l}\times C^{l}}$}} represent the intermediate feature maps. $\mathrm{concat(\cdot,\cdot)}$  represents the channel-wise concatenation operation.

{The output feature map {$\textbf{F}_s\in\mathbb{R}^{H^{l}\times W^{l}\times C^{l}}$} from the {lattice structure} can be obtained as
\begin{equation}
\textbf{F}_s= \textbf{P}_s \oplus \textbf{Q}_s.
\end{equation}}

{Finally, the output feature map           {$\textbf{Y}\in\mathbb{R}^{H^{l}\times W^{l}\times 2C^{l}}$} from LERB is represented as
{\begin{equation}
\textbf{Y} = \mathrm{concat}(\textbf{F}_c, \textbf{F}_s).
\end{equation}}}

{We should point out that there are various ways of combining the contextual module and the spatial module, as shown in Fig.~\ref{fig:LERB-three-different}. In general, the receptive fields of the input feature map are enlarged in the contextual module, while those do not change in the spatial module. As a consequence, when the contextual module and the spatial module are combined as given in Figs.~\ref{fig:LERB-three-different}(b) and \ref{fig:LERB-three-different}(c), the feature maps used for concatenation have different receptive fields.
Such a manner is not only detrimental for feature aggregation, but also increases the learning difficulty of the network. In contrast, the feature maps used for concatenation have the same receptive fields for LERB (i.e., Fig.~\ref{fig:LERB-three-different}(a)). Obviously, this benefits feature aggregation in the decoder. }


Note that {the contextual enhanced block} and {the spatial enhanced block} developed in LERB are different from the basic block that was firstly proposed in \cite{resnet} to address the degradation problem in deep networks. In particular, {the contextual enhanced block} takes advantage of two atrous convolutions (instead of standard convolutions used in the basic block) to enlarge the receptive fields and thus encodes the contextual information. {{The spatial enhanced block} makes use of the downsampled feature map from the first sub-network of ResNet-18 to exploit spatial details.}
{By integrating {the contextual enhanced block} and {the spatial enhanced block} into the lattice structures, LERB effectively enhances both the spatial and contextual information. Moreover, we leverage two {weight learning blocks} to adaptively adjust the weights of two lattice structures. \textit{Such a way generates
various combinations of enhanced blocks, which can enlarge the feature representation space very efficiently}. Hence, compared with the basic block, LERB provides much better feature extraction capability.}

\subsubsection{Feature Transformation Block (FTB)}
\label{approach-FTM}
{For semantic segmentation, it is of great importance to encode both the spatial and {contextual} information for predicting score maps. On the one hand, with the increase of network depth, the high-level feature maps mainly encode the sufficient {contextual} information while lacking spatial details. On the other hand, the low-level feature maps capture the rich spatial information. To exploit multi-level feature maps,
 many modern methods use element-wise addition \cite{RefineNet,Light-weight-refinenet,FCN} or channel-wise concatenation \cite{DFANet,deeplabv3+,U-Net} to aggregate the semantic and spatial feature maps. However, such ways might not be beneficial for semantic segmentation, due to the gap between different levels of feature maps. Therefore, simply aggregating feature maps without taking the differences between feature maps into consideration may not only {cause feature interference, but also decrease the segmentation accuracy.}}

{To address the above problem, {motivated by the Spatial and Channel Squeeze Excitation (scSE) block \cite{scSE}}, we develop FTB to transform the feature map {before} aggregation.  In particular, a transformation tensor is generated to indicate the importance of a feature map, and then is used to weigh each channel and spatial location of a feature map. Therefore, it can be used to emphasize the important information while ignoring the irrelevant information in the input feature map, so that an effective transformed feature map is obtained. In this way, the differences between multi-level feature maps can be greatly alleviated.}

\begin{figure}[t]
\begin{center}
  \includegraphics[width=0.85\linewidth]{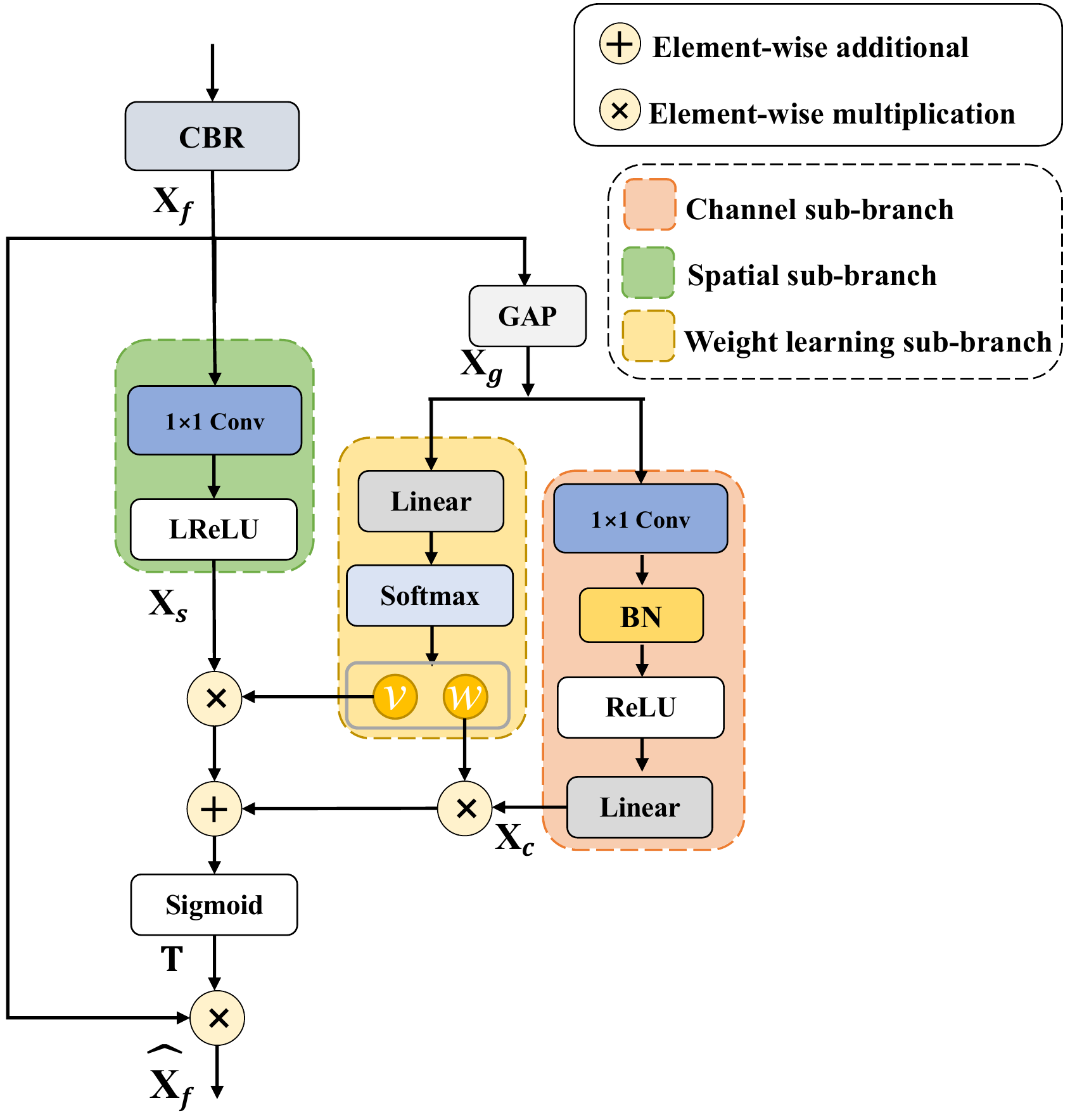}
\end{center}
   \caption{{The network architecture of FTB. {In the figure, ``LReLU" represents the Leaky ReLU activate function.}}}
\label{fig:FTB}
\end{figure}

The network architecture of FTB is shown in Fig. \ref{fig:FTB}. FTB is comprised of two main sub-branches to perform attention operations along the channel and spatial dimensions. Meanwhile, a {weight learning sub-branch} is used to adaptively learn the weights for the channel sub-branch and the spatial sub-branch. {
Roughly, FTB only consists of several convolutional layers and linear operations. Furthermore, the intermediate feature maps in the spatial sub-branch have a small number of channels (i.e., 1) and those in the {channel} sub-branch have a small resolution (i.e., $1 \times 1$). Hence, FTB is a lightweight module.}
%

{More specifically, FTB first employs a {Conv-BN-ReLU} module to generate a feature map $\textbf{X}_{f}\in\mathbb{R}^{H'\times W'\times C'}$. For the spatial sub-branch, the feature map $\textbf{X}_{f}$ is fed into a $1 \times 1$ convolutional layer and {a Leaky ReLU activation function} to obtain the attention tensor $\textbf{X}_{s}\in\mathbb{R}^{H'\times W'\times 1}$. Meanwhile, the feature map $\textbf{X}_{f}$ is also fed into a Global Average Pooling (GAP) layer to obtain the tensor $\textbf{X}_{g}\in\mathbb{R}^{1\times 1\times C'}$ encoding the global information, which can be used in both the channel sub-branch and {weight learning sub-branch}. Then, in the channel sub-branch,  $\textbf{X}_{g}$ is sequentially fed into a $1 \times 1$ convolutional layer, a BN layer, a ReLU activation function and a linear layer to obtain the attention tensor $\textbf{X}_{c}\in\mathbb{R}^{1\times 1\times C'}$.  In the {weight learning sub-branch}, $\textbf{X}_{g}$ is fed into a linear layer followed by a softmax activation function to adaptively learn two weights $v$ and $w$. Therefore, the transformation tensor $\textbf{T}\in\mathbb{R}^{H'\times W'\times C'}$ can be computed as
\begin{equation}
{\textbf{T} = \beta(\eta(v\textbf{X}_{s}) + \eta(w\textbf{X}_{c})),}
\end{equation}
{where $\beta(\cdot)$ denotes the Sigmoid activation function.
 Before the addition operation, the channel attention values are broadcast (copied) along the spatial dimension, while the spatial attention values are broadcast along the channel dimension. }
}


%
%

{Finally, the element-wise multiplication is performed between $\textbf{X}_{f}$ and $\textbf{T}$, which is formulated as}

{
\begin{equation}
    \widehat{\textbf{X}}_{f} = \textbf{T} \otimes \textbf{X}_{f},
\end{equation}
where $\widehat{\textbf{X}}_{f}$ represents the transformed feature map.}

{It is worth noting that
the scSE block also learns the attention on both spatial and channel dimensions. However, different from the scSE block, the proposed FTB not only applies the attention sub-branches on both spatial and channel dimensions to obtain attention tensors, but also adaptively learns weights by designing a {weight learning sub-branch}. Hence, the attention tensors from the spatial and channel sub-branches can be effectively combined to obtain a transformation tensor. Furthermore, the scSE block uses the squeeze operator and the excitation operator on the channel dimension. However, the squeeze operator may lead to information loss since the channel number is reduced. In contrast, FTB removes these operators to model the dependencies between channels more accurately.}



\subsubsection{Global Context Block (GCB)}
\label{approach-GCM}
Currently, most semantic segmentation methods are based on the DCNN models that are originally designed for the image classification task. Such a task relies largely  on the high-level semantic information (such as object-level or category-level evidence). 
These DCNN models, however, may not accurately identify and locate the objects due to the lack of global contextual information, thus leading to a negative impact on the accuracy of semantic segmentation. Therefore, the global contextual information plays a critical role in street scene segmentation.

Based on the above observations, similar to BiSeNet \cite{BiSeNet},
we append a GCB at the end of ResNet-18 to exploit the contextual information of the image.
The network architecture of GCB is shown in Fig. \ref{fig:architecture-network}. {GCB first performs the GAP operation on  {the} feature map (whose size is $1/32$ of the original {input image size}) from sub-network 4 of ResNet-18 to obtain a $1 \times 1$  feature map with the largest receptive fields.}
{Then, the feature map is passed through a {Conv-BN-ReLU} module. Finally, the bilinear interpolation is used to restore the feature map back to $1/32$ of the original input image size.}
In fact, compared with the pooling features with multiple window sizes used in RefineNet \cite{RefineNet}, GCB has smaller memory consumption and less floating-point operations.




\subsection{Joint Loss}
\label{loss}
In DMA-Net, both the auxiliary loss and principal loss are employed to optimize the training of the network. In particular, the auxiliary losses are used to supervise the training of the MB and HB of MAN, and the principal loss is employed to supervise the output of the whole network (i.e., the output from the LB of MAN). To be specific, the joint loss is formulated as
{
\begin{equation}\label{pa}
\begin{split}
\mathcal{L}_{joint} =& \mathcal{L}_{principal}(\textbf{O}^p,\textbf{O})
+ \lambda[\mathcal{L}_{auxiliary}(\textbf{O}^p_{mid},\textbf{O}) \\
& +  \mathcal{L}_{auxiliary}(\textbf{O}^p_{high},\textbf{O})],
\end{split}
\end{equation}}
\noindent where $\mathcal{L}_{joint}$, $\mathcal{L}_{principal}$, and $\mathcal{L}_{auxiliary}$ represent the joint loss, the principal loss, and the auxiliary loss, respectively. {$\lambda$ denotes the balance weight.
$\textbf{O}^p$ denotes the predicted output from the whole network.  $\textbf{O}^p_{mid}$ and $\textbf{O}^p_{high}$ denote the resized predicted outputs (having the same size as the input image) from the MB and HB of MAN, respectively.
$\textbf{O}$ denotes the ground-truth semantic labels.}

All the loss functions adopt the pixel-wise cross entropy, whose form is defined as follows:
\begin{equation}\label{ce}
\mathcal{L}(\textbf{Z}^p;\textbf{Z}) = -\frac{1}{N}\sum_{i}^{K}\sum_{j}^{N}z_{i,j}\log(z^p_{i,j}),
\end{equation}
where $\textbf{Z}^p$ is the predicted output given by the softmax function and $\textbf{Z}$ is the ground-truth semantic labels. $z^p_{i,j}$ and $z_{i,j}$ denote the probability value of the $i$-th category at the $j$-th pixel location of the output and its corresponding ground-truth label, respectively. $N$ is the total number of pixels and $K$ is the total number of semantic categories.

\subsection{{Discussions}}
\label{discuss}
{Both our proposed  DMA-Net method and some recent real-time semantic segmentation methods \cite{pre-work, BiSeNet, SwiftNet} take advantage of the encoder-decoder structure to improve the segmentation accuracy. However, there are significant differences between DMA-Net and these methods.}

{First, we propose LERB to address the problem of inferior feature extraction capability of the lightweight backbone network. Specifically, LERB enhances the spatial detail and context information in the feature maps by two feature enhancement blocks (i.e., a contextual enhanced block and a spatial enhanced block). In particular, LERB can expand the representation space of features by introducing the lattice structures. Hence, LERB effectively and efficiently improves the feature representations of the network.
 In contrast,  previous methods either use additional branches for feature enhancement (such as BiSeNet \cite{BiSeNet}, RTHP\cite{pre-work}), or employ a parallel network structure to enlarge the receptive fields of the network (such as SwiftNet \cite{SwiftNet}). Although these methods can enhance feature maps to a certain extent, additional branches or parallel networks will also bring high computational costs.} 

{Second, we leverage FTB to reduce the gap between different levels of feature maps. Specifically, we {use a weight learning sub-branch} in FTB to adaptively enhance the important information and suppress the irrelevant information. Therefore, the problem of feature interference between different levels of feature maps is greatly alleviated, so that the spatial and contextual information in these feature maps can be properly aggregated. On the contrary, many methods \cite{pre-work,DFANet} adopt simple aggregation operations (such as the element-wise addition and the channel-wise concatenation) to aggregate different levels of feature maps. Hence, they ignore the differences between feature maps, resulting in a performance decrease.}

\section{Experiments}
\label{sec:experiments}
In this section, we evaluate the performance of the proposed DMA-Net on the challenging street scene benchmarks (including the Cityscapes and {CamVid} datasets). We first introduce the datasets and evaluation metrics in Section \ref{Exp:datasets}.  Then, we describe the implementation details in Section \ref{Exp:detail}. Next, we conduct ablation studies to analyze the effectiveness of each key component of {DMA-Net} in Section \ref{Exp:ablation-studies}. We compare {DMA-Net} with several state-of-the-art real-time semantic segmentation methods on the
Cityscapes and CamVid datasets
in Section \ref{Exp:compare} and Section \ref{Exp:compare-camvid}, respectively. {Finally, we discuss the limitations of {DMA-Net} in Section \ref{Limitation}.}

\subsection{Datasets and Evaluation Metrics}
\label{Exp:datasets}

The Cityscapes dataset consists of 25,000 high-resolution (with the size of $1024 \times 2048$) street scene images that were collected from 50 different cities in Germany. These images are divided into two parts: 5,000 fine-annotated images and 20,000 weakly-annotated images. In this paper, we only use the fine-annotated images in our experiments. These fine-annotated images can be classified into 30 categories and split into three  datasets: a training  dataset (including 2,975 images), a validation dataset (including 500 images), and a test  dataset (including 1,525 images). Similar to state-of-the-art semantic segmentation methods \cite{erfnet,ICNet}, we only use 19 common semantic categories (such as sidewalk, road, and car) in our experiments. For the test  dataset, we evaluate our method by using the online service provided by Cityscapes, which do not release the ground-truth images to users.

The CamVid dataset is another challenging semantic dataset for street scene understanding. It consists of 701 high-resolution (with the size of $720 \times 960$) video frames collected from five video sequences and 11 semantic categories. For a fair comparison, we split the whole dataset into training, validation, and test datasets, which respectively contain 367 images, 101 images, and 233 images, as done in \cite{SegNet}.

For evaluation metrics, we adopt mean Intersection over Union (mIoU) and  {Frames Per Second (FPS)}, which measure the segmentation accuracy and latency, respectively.  Moreover, we also use the number of parameters (Params) and floating-point operations (FLOPs) to evaluate the memory consumption and computational complexity of the model, respectively.

\subsection{Implementation Details}
\label{Exp:detail}

For training, we employ the horizontal flipping, random scaling (the scale ratio ranges from 0.5 to 2.0), and random cropping on all the images to augment the dataset. {The final image resolution for Cityscapes is $768 \times 1536$ and that for CamVid is $640 \times 640$.} 
All the network parameters of the convolutional layers in ResNet-18 are initialized from the publicly available ResNet-18 \cite{resnet} pretrained on the ImageNet \cite{ImageNet}. The network parameters of MAN and GCB are {randomly initialized by using the Kaiming normal initialization\cite{kaiming-init}.}

{To optimize the whole network, we adopt Stochastic Gradient Descent (SGD) \cite{sgd} with the batch size of 16, the momentum of 0.9, and the weight decay of 0.0005 to update the network parameters for Cityscapes.} Moreover, we utilize the online hard example mining \cite{ohem} to mitigate the influence of class imbalance. {Similar to state-of-the-art semantic segmentation methods, we use the popular "poly" learning rate strategy $(1-\frac{iter}{total\_iters})^{power}$ with the power of 0.9 to update the learning rate, where the initial learning rate is set to 0.005. For CamVid, the batch size and learning rate are set to 4 and 0.001, respectively.}

{The whole training process contains 60,000 iterations for Cityscapes and 80,000 iterations for CamVid. Codes are implemented by the PyTorch framework.
All experiments on speed analysis are performed by using a single NVIDIA GTX 1080Ti GPU.}

\begin{table}
    \caption{{The Accuracy , Speed, and Params Analysis of Different Backbone Networks: MobileNetV2, ResNet-101, and ResNet-18 on the Cityscapes Validation Dataset.}}
    \begin{center}
    \setlength{\tabcolsep}{1.8mm}{
    \renewcommand{\arraystretch}{1.3}
    \begin{tabular}{l|ccc}
        \whline
        Backbone Network & mIoU (\%) & Speed (FPS) & Params (M) \\
        \hline\hline
        FCN+MobileNetV2 & 61.7 & 28 & \textbf{2.04} \\
        FCN+ResNet-101 & \textbf{65.2} & 9 & 51.95 \\
        {FCN+ResNet-50} & {64.1} & {19} & {32.95} \\
        FCN+ResNet-18 & 63.6 & \textbf{54} & 11.77\\
        \whline
    \end{tabular}
    }
    \end{center}
    \label{Exp:ResNet-18}
\end{table}

\subsection{Ablation Studies}
\label{Exp:ablation-studies}
In this subsection, we investigate the effectiveness of each key component of {DMA-Net} (including ResNet-18, {MAN, LERB, FTB, and GCB)} step-by-step. In the following experiments, we evaluate these components on the Cityscapes validation dataset \cite{cityscapes}.



\subsubsection{{Effectiveness of ResNet-18}}
{In this paper, we employ ResNet-18 (a lightweight version of ResNet) as our backbone network (the encoder of DMA-Net).  As we mentioned above, the backbone network provides the basic feature extraction for the whole network, and it can affect both the segmentation
accuracy and the inference speed of semantic segmentation. Complicated backbone networks have a large number of network parameters and floating-point operations, leading to serious degradation of the inference speed. Therefore, lightweight
networks are usually adopted as backbone networks for real-time semantic segmentation.}

{To evaluate the effectiveness and efficiency of ResNet-18, we compare it with {three widely used backbone networks (including MobileNetV2 \cite{mobilenetv2}, ResNet-50 \cite{resnet}, and ResNet-101 \cite{resnet}).} For simplicity, all the backbone networks are pretrained on the ImageNet dataset and use FCN \cite{FCN} as the base structure. The comparison results are
shown in Table \ref{Exp:ResNet-18}.}




We can see that FCN+ResNet-101 achieves the highest segmentation accuracy (about 65.2\% mIoU), which is about 3.5\% and 1.6\% higher than FCN+MobileNetV2 and FCN+ResNet-18, respectively. {However, the number of network parameters of FCN+ResNet-101 is significantly high (about 51.95M), and its inference speed is the slowest (about 9 FPS) among all the competing methods.} Although FCN+MobileNetV2 has the smaller  number of parameters   than the other {three} backbone networks, it achieves the lowest mIoU. {FCN+ResNet-50 achieves 64.1\% mIoU in terms of segmentation accuracy and the inference speed of 19 FPS.}
Note that FCN+ResNet-18 achieves worse segmentation accuracy than {FCN+ResNet-101 and FCN+ResNet-50}, but its number of parameters is much smaller. {Moreover, FCN+ResNet-18 has much faster inference speed than the other competing methods.} {This shows that ResNet-18 can achieve a good balance between accuracy and inference speed. In the following, we will fix ResNet-18 as our encoder.}




\begin{figure*}[!tp]
\begin{center}
\includegraphics[width=0.9\linewidth]{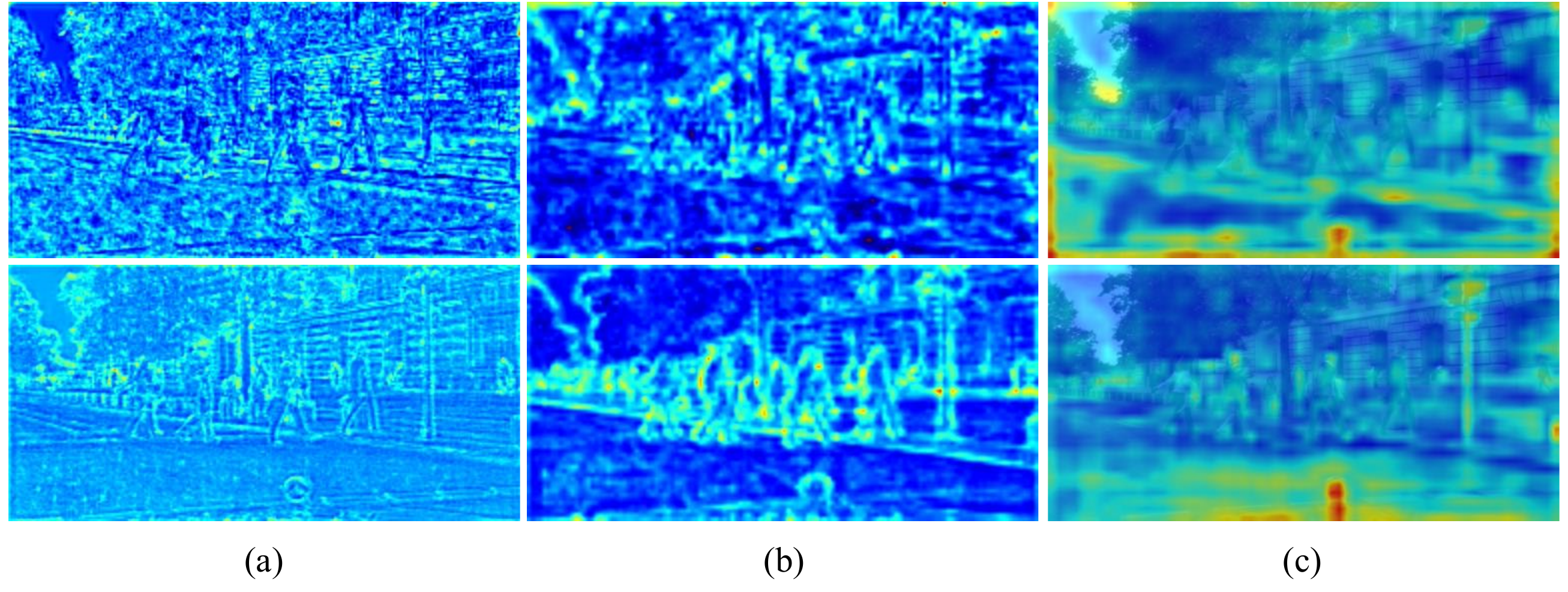}
\end{center}
  \caption{{Visualization results of feature maps. The images from the first column to the last column represent the feature maps from the (a) LB, (b) MB, and (c) HB, respectively. The upper panel and the lower panel show the feature maps before and after LERB, respectively.}}
\label{fig:visual-result}
\end{figure*}

\begin{table}
    \caption{{The Influence of GCB, LERB, and FTB on the Cityscapes Validation Dataset.}}
    \begin{center}
    \setlength{\tabcolsep}{1.5mm}{
    \renewcommand{\arraystretch}{1.3}
    \begin{tabular}{l|ccc}
        \whline
        Method & mIoU (\%) & Speed (FPS) & Params (M)\\
        \hline\hline
        {Baseline} & {72.9} & {55.3} & {12.99} \\
        {Baseline+GCB} & {74.1} & {55.0} & {13.13} \\
        {Baseline+GCB+FTB} & {75.7} & {54.4} & {13.50} \\
        {Baseline+GCB+LERB} & {76.3} & {47.4} & {14.23} \\
        {DMA-Net} & {76.8} & {46.7} & {14.60} \\
        \whline
    \end{tabular}
    }
    \end{center}
    \label{exp:MAN}
\end{table}

\subsubsection{{Effectiveness of MAN}}
{To demonstrate the effectiveness of MAN (the decoder of DMA-Net), we evaluate the influence of different combinations of key components on the accuracy, speed, and memory consumption, as shown in Table \ref{exp:MAN}.
The Baseline method adopts the encoder-decoder structure and it consists of ResNet-18 and a simplified version of MAN, where LERB, FTB, and GCB are not used. The Baseline+GCB, Baseline+GCB+FTB, Baseline+GCB+LERB, and DMA-Net methods share the same network architectures as the Baseline method, except that GCB, GCB+FTB, GCB+LERB, and GCB+LERB+FTB are respectively employed in MAN.}


{By comparing Table \ref{Exp:ResNet-18} and Table \ref{exp:MAN}, the Baseline method achieves 72.9\% mIoU, which is much higher than FCN+ResNet-18 (about 9.3\% mIoU {higher}). This demonstrates the superiority of the encoder-decoder structure. Compared with the Baseline method, the Baseline+GCB method achieves better segmentation accuracy (about 1.2\% mIoU higher). This result validates the importance of GCB.}



{Incorporating {FTB or LERB} into the Baseline+GCB method can further boost the segmentation accuracy. By taking into account both LERB and FTB, our DMA-Net method achieves the highest mIoU (about {76.8}\%).} The above results show that both LERB and FTB are beneficial to improve the performance of semantic segmentation. This is because the joint learning of LERB and FTB enables the network to effectively aggregate hierarchical feature maps.

{From the perspective of inference speed, the speed of Baseline+GCB is only slightly slower than that of Baseline. Hence, GCB brings only a small computational cost. By combining LERB with Baseline+GCB, the speed of the Baseline+GCB+LERB method is only about {7.6} FPS slower than that of the Baseline+GCB method. This shows the efficiency of LERB.
 Meanwhile, FTB has a subtle influence on the inference speed, since the speed of the Baseline+GCB+FTB method is only slightly slower than that of the Baseline+GCB method. Similarly, the inference speed of DMA-Net is almost the same as that of Baseline+GCB+LERB.}
 {In terms of the number of network parameters, the differences between all the competing methods are not significant. Thus, the memory consumption of these methods is relatively small ($<15$M).}

 {In summary, the above experimental results show that by incorporating GCB, LERB, and FTB into MAN, our method is able to achieve a good tradeoff between speed and accuracy.}

\begin{table}[t]
    \caption{{The Accuracy, Speed, and Params Comparison Between LERB, LERB-addition, and Residual Building Blocks: Basicblock, Bottleneck on the Cityscapes Validation Dataset.}}
    \begin{center}
    \setlength{\tabcolsep}{2mm}{
    \renewcommand{\arraystretch}{1.3}
    \begin{tabular}{l|ccc}
        \whline
        Method & mIoU (\%) & Speed (FPS) & Params (M) \\
        \hline\hline
        {DMA-Net (Basicblock)} & {75.8} & {49.7} & {15.05} \\
       {DMA-Net (Bottleneck)} & {75.6} & {50.9} & {13.59} \\
       \hline
        {DMA-Net (LERB-addition)} & {76.1} & {47.9} & {14.60} \\
        {DMA-Net (LERB-b)} & {76.4} & {46.7} & {14.60} \\
        {DMA-Net (LERB-c)} & {76.3} & {46.7} & {14.60} \\
        {DMA-Net}  & {76.8} & {46.7} & {14.60} \\
        \hline

        \whline
    \end{tabular}
    }
    \end{center}
    \label{Exp:LERB}
\end{table}

\subsubsection{Effectiveness of LERB}
In this subsection, we evaluate the effectiveness of our proposed LERB. We replace the LERB in DMA-Net with the Basicblock and Bottleneck used in ResNet \cite{resnet}, respectively. The comparison results are shown in Table \ref{Exp:LERB}.

{We can observe that the mIoU obtained by DMA-Net is improved by about 1.0\% in comparison with DMA-Net (Basicblock). Moreover, compared with DMA-Net (Bottleneck), DMA-Net  also achieves higher accuracy (about 1.2\% mIoU higher). With regards to speed, DMA-Net is only about 3 FPS and 4.2 FPS slower than DMA-Net (Basicblock) and DMA-Net (Bottleneck), respectively.}
These results demonstrate that LERB can {enhance feature representations of our network more effectively than the other residual blocks for real-time semantic segmentation in street scenes.}

In order to further investigate the effectiveness of {the lattice structure} in LERB on the final performance, we also replace {the lattice structure} in LERB with the element-wise addition operation, named DMA-Net (LERB-addition). As we can see, compared with the simple element-wise addition operation, adopting {the lattice structure} in LERB  improves the segmentation accuracy by about 0.7\% mIoU with a slight drop in terms of inference speed. This indicates that feature maps with different  combinations in {the lattice structure} can effectively improve the representation capability of the network in an efficient manner.

{Then, we compare LERB with its two variants.
The two variants are denoted as DMA-Net (LERB-b) and DMA-Net (LERB-c) according to the structures given in Figs.~\ref{fig:LERB-three-different}(b) and \ref{fig:LERB-three-different}(c), respectively. We can see that the accuracy obtained by DMA-Net (LERB-b) and DMA-Net (LERB-c) is lower than that obtained by DMA-Net. This is because the feature maps used for concatenation have different receptive fields, which have an adverse effect on the feature aggregation, thereby reducing the final segmentation performance.}

{Finally, we give some visualization results to show the importance of LERB, as shown in Fig.~\ref{fig:visual-result}. To be specific, we visualize the feature maps before and after LERB in three branches of MAN for DMA-Net. We can observe that LERB enables the network to enhance spatial details and context information of feature maps in three branches. For example, the feature map after LERB pays more attention to edge details in the LB, while it focuses more on the context information in the HB.}

\begin{table}[t]
    \caption{{The Accuracy, Speed, and Params Comparison Between Different Transformation Blocks on the Cityscapes Validation Dataset.}}
    \begin{center}
    \setlength{\tabcolsep}{2mm}{
    \renewcommand{\arraystretch}{1.3}
    \begin{tabular}{l|ccc}
        \whline
        Method & mIoU (\%) & Speed (FPS) & Params (M) \\
        \hline\hline
        {DMA-Net\ (scSE)} & {76.5} & {46.6} & {14.25} \\
        \hline
        {DMA-Net\ (FTB\_WLB)} & {76.4} & {46.8} & {14.60} \\
        {DMA-Net} & {76.8} & {46.7} & {14.60} \\
        \whline
    \end{tabular}
    }
    \end{center}
    \label{Exp:FTB}
\end{table}

\subsubsection{Effectiveness of FTB}
In this subsection, we further study the importance of FTB.
The results are listed in Table \ref{Exp:FTB}.
{{DMA-Net (scSE) denotes the method that has the same network architecture as DMA-Net, except that FTB is replaced with the scSE block \cite{scSE}}.}

\begin{figure}[!tp]
\begin{center}
\includegraphics[width=1.0\linewidth]{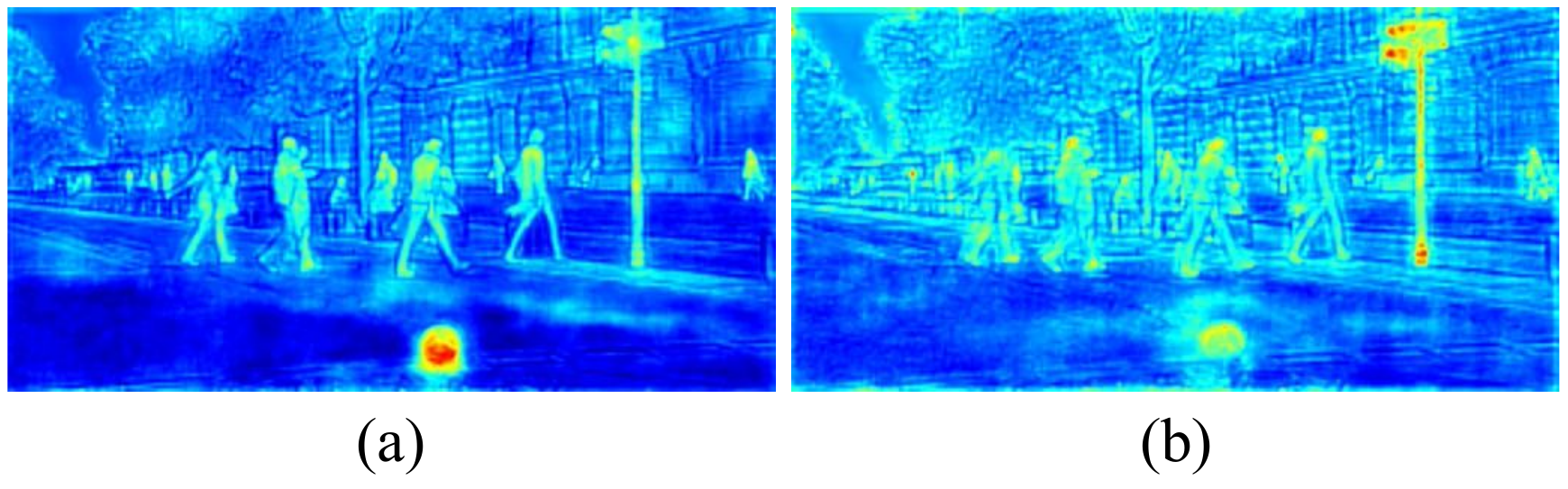}
\end{center}
  \caption{{Visualization results of feature maps obtained by (a) Baseline+GCB and (b) Baseline+GCB+FTB, respectively.}}
\label{fig:visual-result-ftb}
\end{figure}

From Table \ref{Exp:FTB}, we can see that the segmentation accuracy obtained by DMA-Net is higher than DMA-Net\_scSE (about {0.3}\% mIoU improvement). This is because FTB adaptively combines the spatial and channel sub-branches with a weight learning sub-branch. Moreover, FTB preserves more information than scSE in the channel dimension (note that the  squeeze operator used in scSE is not adopted in FTB).
Therefore, FTB is able to obtain informative transformed feature maps.
As a result, different levels of feature maps can be effectively aggregated. {From the perspective of speed, DMA-Net and DMA-Net (scSE) obtain almost the same inference speed. In terms of network parameters, DMA-Net is only slightly higher than DMA-Net (scSE).}

{We further investigate the effect of adaptive weights on the final segmentation performance. We denote DMA-Net without using the {weight learning sub-branch} in FTB as DMA-Net (FTB\_WLB). DMA-Net achieves better accuracy than DMA-Net (FTB\_WLB). This shows the importance of {the weight learning sub-branch} in FTB.}

{Finally, we also visualize the feature maps (the output of MAN) obtained by Baseline+GCB and Baseline+GCB+FTB, respectively. Some visualization results are shown in Fig.~\ref{fig:visual-result-ftb}. Compared with the feature map obtained by Baseline+GCB, the feature map obtained by Baseline+GCB+FTB not only preserves finer edge details, but also better focuses on objects at different scales. This validates that FTB can effectively reduce the gap between different levels of feature maps and thus facilitates the combination of spatial details and semantic information. }


\subsubsection{Effectiveness of GCB}

{In this subsection, we further verify the effectiveness of GCB. We compare GCB with GAP. The comparison results are as shown in Table \ref{Exp:GCB}, where DMA-Net (GAP) shares the same network {architecture} as DMA-Net, except that GCB is replaced with GAP.}



{Compared with the DMA-Net (GAP) method, the DMA-Net method increases about 0.3\% mIoU, which indicates the superiority of GCB. GCB can effectively capture the global contextual information. This is due to the fact that we use the convolutional operation after the GAP operation, which enables the network to extract more compact global feature representations, thus improving the final segmentation performance. Meanwhile, GCB has little influence on the inference speed, since DMA-Net is only slightly slower than DMA-Net (GAP).}

%


\subsubsection{{Influence of Auxiliary Loss}}
\begin{figure}[!tp]
\begin{center}
\includegraphics[width=0.75\linewidth]{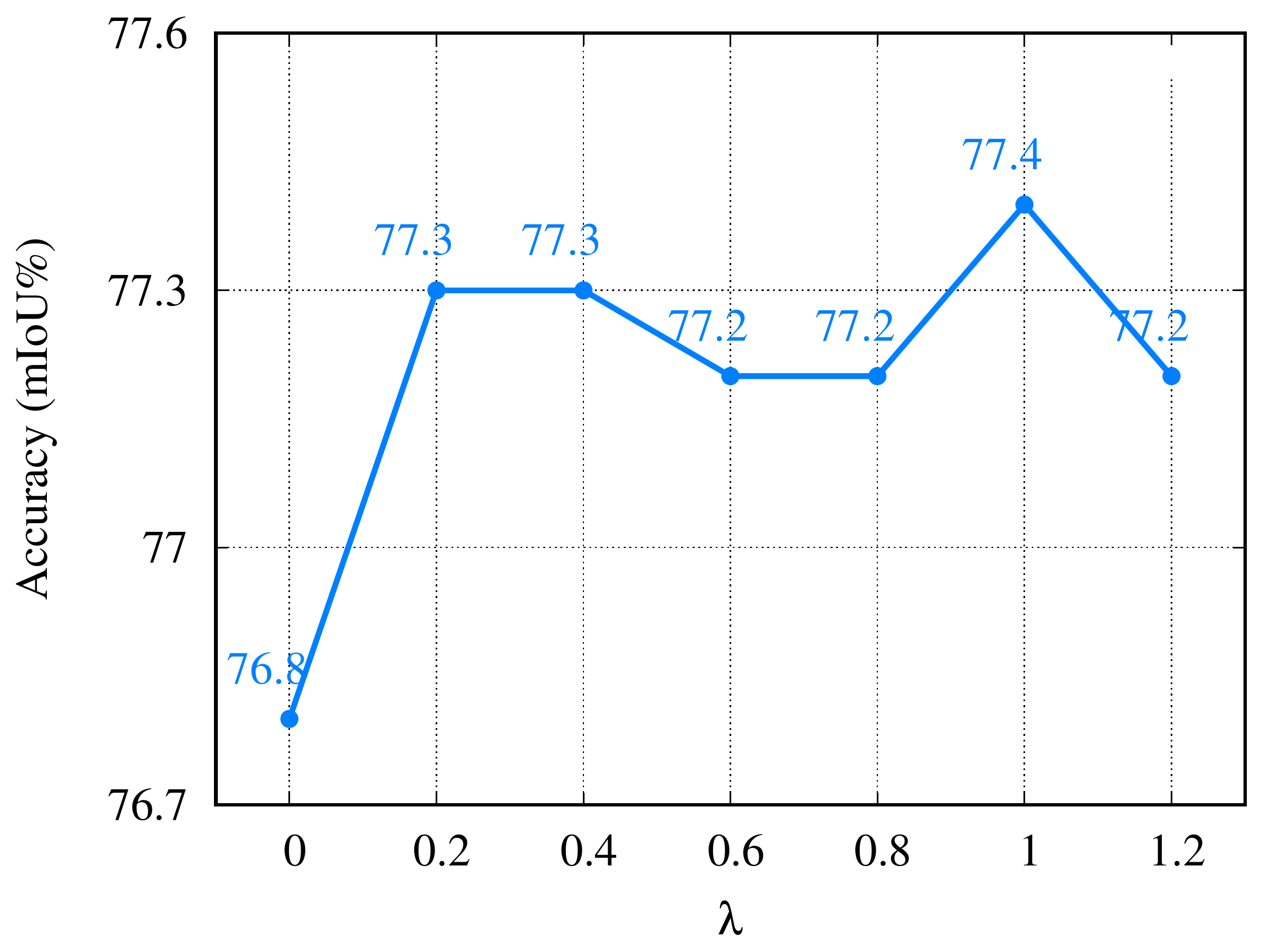}
\end{center}
  \caption{{The accuracy obtained by the proposed DMA-Net with different values of the parameter $\lambda$ on the Cityscapes validation dataset.}}
\label{fig:aux}
\end{figure}

{In this subsection, we evaluate the influence of auxiliary loss on the final performance. In the experiments, we change the balance weight $\lambda$ from 0 to 1.2. All the results are shown in Fig.~\ref{fig:aux}.}

{In Fig.~\ref{fig:aux}, we can observe that the accuracy obtained by DMA-Net is only slightly different when the values of $\lambda$ are within the range of  $[0.2, 1.2]$.
This shows that the network is not very sensitive to the value of $\lambda$. When $\lambda=1$, our proposed method achieves the best performance ({77.4}\% mIoU). Therefore, employing the auxiliary loss in MAN is beneficial to improve the segmentation performance. When $\lambda=0$, only the principal loss is used to supervise the training. In this case, the mIoU obtained by our method drops to {76.8}\%. The above results show that the proposed auxiliary loss enables our method to explicitly supervise the training of the MB and HB of MAN, thus improving the segmentation performance.}

\begin{table}[!t]
    \caption{{The Accuracy, Speed, and Params Comparison Between Different Global Contextual Modules on the Cityscapes Validation Dataset.}}
    \begin{center}
    \setlength{\tabcolsep}{2.8mm}{
    \renewcommand{\arraystretch}{1.3}
    \begin{tabular}{l|ccc}
        \whline
        Method & mIoU (\%) & Speed (FPS) & Params (M) \\
        \hline\hline
        {DMA-Net (GAP)} & {76.5} & {46.8} & {14.60} \\
        DMA-Net & {76.8} & {46.7} & {14.60} \\
        \whline
    \end{tabular}
    }
    \end{center}
    \label{Exp:GCB}
\end{table}

\subsection{Comparisons with State-of-the-Art Methods}
\label{Exp:compare}

To evaluate the effectiveness and efficiency of {DMA-Net, we first compare it with the simplified PSPNet \cite{PSPNet} and SwiftNet \cite{SwiftNet} on the Cityscapes validation dataset. The simplified PSPNet is obtained by compressing the kernel keeping rate of PSPNet and SwiftNet is the current representative real-time semantic segmentation method. All the results are reported in Table \ref{table:compare-one}.}

\begin{table}[!t]
    \caption{{The Accuracy and Speed Comparison Between The Proposed Method and PSPNet, ICNet on the Cityscapes Validation Dataset.}}
    \begin{center}
    \setlength{\tabcolsep}{2.5mm}{
    \renewcommand{\arraystretch}{1.1}
    \begin{tabular}{l|cccc}
        \whline
        Item & PSPNet & SwiftNet & DMA-Net (ours) \\
        \hline\hline
        mIoU (\%) & 67.9 & 75.4 & {\textbf{77.4}} \\
        Time (ms) & 170 & 25 & \textbf{21.4} \\
        Speed (FPS) & 5.9 & 39.9 & \textbf{46.7} \\
        Image size & $713\times713$ & $1024\times2048$ & $1024\times2048$  \\
        \whline
    \end{tabular}
    \label{table:compare-one}
    }
    \end{center}
\end{table}

It can be seen that {DMA-Net outperforms the two competing methods and achieves an overwhelming performance with {77.4}\% mIoU accuracy at the inference speed of 46.7 FPS.} Specifically, the segmentation accuracy of our proposed DMA-Net exceeds that of the simplified PSPNet and ICNet by about {10}\% and 2\%, respectively. Meanwhile, the inference speed of our method is much faster than that of the two competing methods. The results demonstrate that DMA-Net provides excellent inference speed and high segmentation accuracy on the Cityscapes validation dataset.

\begin{table*}[!tp]
    \caption{Comparisons Between the Proposed Method and Other State-of-the-art Methods on the Cityscapes Test Dataset. ``-" Indicates That the Corresponding Result is Not Provided by the Method.}
    \begin{center}
    \setlength{\tabcolsep}{5mm}{
    \renewcommand{\arraystretch}{1}
    \begin{tabular}{l|c|c|c|c|c}
        \whline
        Method & Input Size & FLOPs (G) & Params (M) & Speed (FPS) & mIoU (\%)  \\
        \hline\hline
        DeepLab \cite{deeplab} & $512\times1024$ & 457.8 & 262.1 & 0.25 & 63.1 \\
        PSPNet \cite{PSPNet} & $713\times713$ & 412.2 & 250.8 & 0.78 & 78.4 \\
        \hline
        SegNet \cite{SegNet} & $640\times360$ & 286 & 29.5 & 14.6 & 56.1 \\
        ENet \cite{ENet} & $630\times630$ & 4.4 & 0.4 & 76.9 & 58.3 \\
        ESPNet \cite{espnet} & $512\times1024$ & 4.7 & 0.4 & 112 & 60.3 \\
        SQNet \cite{SQNet} & $1024\times2048$ & 270 & - & 16.7 & 59.8 \\
        CRF-RNN \cite{CRF-RNN} & $512\times1024$ & - & - & 1.4 & 62.5 \\
        FCN-8S \cite{FCN} & $512\times1024$ & 136.2 & - & 2.0 & 65.3 \\
        FRRN \cite{FRRN} & $512\times1024$ & 235 & - & 2.1 & 71.8 \\
        ERFNet \cite{erfnet} & $512\times1024$ & - & 2.1 & 41.7 & 68.0 \\
        ICNet \cite{ICNet} & $1024\times2048$ & 29.8 & 26.5 & 30.3 & 69.5 \\
        TwoColumn \cite{TwoColumn} & $512\times1024$ & 57.2 & - & 14.7 & 72.9 \\
        DFANet \cite{DFANet} & $1024\times1024$ & 3.4 & 7.8 & 100.0 & 71.3 \\
        LEDNet \cite{LEDNet} & $512\times1024$ & - & 0.94 & 71 & 70.6 \\
        RTHP \cite{pre-work} & $448\times896$ & 49.5 & 6.2 & 51.0 & 73.6 \\
        BiSeNet1 \cite{BiSeNet} & $768\times1536$ & 14.8 & 5.8 & 72.3 & 68.4 \\
        BiSeNet2 \cite{BiSeNet} & $768\times1536$ & 55.3 & 49 & 45.7 & 74.7 \\
        SwiftNet \cite{SwiftNet} & $1024\times2048$ & 104 & 11.8 & 39.9 & 75.5 \\
        \hline
        \hline
        {DMA-Net (small)} & {$768\times1536$} & {53.0} & {14.60} & {76.8} & {\textbf{75.6}} \\
        DMA-Net & $1024\times2048$ & {94.2} & {14.60} & {46.7} & {\textbf{77.0}} \\
        \whline
    \end{tabular}
    \label{table:compare-two}
    }
    \end{center}
\end{table*}
\begin{table*}[!tp]
    \caption{The Per-class, Class and Category IoU(\%) on the Cityscapes Test Dataset for DMA-Net Compared to Other Methods. List of Classes (from Left to Right): Road, Side-walk, Building, Wall, Fence, Pole, Traffic Light, Traffic Sign, Vegetation, Terrain, Sky, Pedestrian, Rider, Car, Truck, Bus, Train, Motorbike and Bicycle. ``Cla" Denotes Miou (19 Classes),  {``Var" Denotes the Variance. }}
    \begin{center}
    \setlength{\tabcolsep}{1.2mm}{
    \begin{tabular}{l||ccccccccccccccccccc||cc}
        \whline
        Method & Roa & Sid & Bui & Wal & Fen & Pol & TLi & TSi & Veg & Ter & Sky & Ped & Rid & Car & Tru & Bus & Tra & Mot & Bic & Cla & Var \\
        \hline\hline
        SegNet \cite{SegNet} & 96.4 & 73.2 & 84.0 & 28.4 & 29.0 & 35.7 & 39.8 & 45.1 & 87.0 & 63.8 & 91.8 & 62.8 & 42.8 & 89.3 & 38.1 & 43.1 & 44.1 & 35.8 & 51.9 & 57.0 & {5.09} \\
        ENet \cite{ENet} & 96.3 & 74.2 & 75.0 & 32.2 & 33.2 & 43.4 & 34.1 & 44.0 & 88.6 & 61.4 & 90.6 & 65.5 & 38.4 & 90.6 & 36.9 & 50.5 & 48.1 & 38.8 & 55.4 & 58.3 &  {4.61} \\
        FCN-8s \cite{FCN} & 97.4 & 78.4 & 89.2 & 34.9 & 44.2 & 47.4 & 60.1 & 65.0 & 91.4 & 69.3 & 93.9 & 77.1 & 51.4 & 92.6 & 35.3 & 48.6 & 46.5 & 51.6 & 66.8 & 65.3 &  {4.11} \\
        ERFNet \cite{erfnet} & 97.9 & 82.1 & 90.7 & 45.2 & 50.4 & 59.0 & 62.6 & 68.4 & 91.9 & 69.4 & 94.2 & 78.5 & 59.8 & 93.4 & 52.3 & 60.8 & 53.7 & 49.9 & 64.2 & 69.7 &  {2.85} \\
        LEDNet \cite{LEDNet} & 98.1 & 79.5 & 91.6 & 47.7 & 49.9 & 62.8 & 61.3 & 72.8 & 92.6 & 61.2 & 94.9 & 76.2 & 53.7 & 90.9 & 64.4 & 64.0 & 52.7 & 44.4 & 71.6 & 70.6 &  {2.82}  \\
        {{BiSeNet2} \cite{BiSeNet}} & {98.2} & {82.9} & {91.7} & {44.5} & {51.1} & {\textbf{63.5}} & {\textbf{71.2}} & {75.0} & {92.9} & {71.1} & {94.9} & {83.6} & {65.4} & {94.9} & {60.5} & {68.7} & {56.8} & {61.5} & {72.7} & {73.8} & {2.40}  \\
        {SwiftNet \cite{SwiftNet}} & {98.3} & {83.9} & {92.2} & {46.3} & {52.8} & {63.2} & {70.6} & {\textbf{75.8}} & {\textbf{93.1}} & {70.3} & {\textbf{95.4}} & {84.0} & {64.5} & {95.3} & {63.9} & {78.0} & {71.9} & {61.6} & {\textbf{73.6}} & {75.5} & {2.15}  \\
        \hline\hline
        {DMA-Net} & {\textbf{98.5}} & {\textbf{85.5}} & {\textbf{92.2}} & {\textbf{53.3}} & {\textbf{55.3}} & {62.5} & {70.9} & {74.9} & {93.0} & {\textbf{71.2}} & {95.0} & {\textbf{84.0}} & {\textbf{66.6}} & {\textbf{95.6}} & {\textbf{68.2}} & {\textbf{82.8}} & {\textbf{76.6}} & {\textbf{64.5}} & {73.2} & {\textbf{77.0}} & {\textbf{1.82}} \\
        \whline
    \end{tabular}
    \label{table:compare-two-per-class}
    }
    \end{center}
\end{table*}

Then, we compare our proposed method with several state-of-the-art real-time semantic segmentation methods on the Cityscapes test dataset, as given in Table \ref{table:compare-two}. In Table \ref{table:compare-two},  the inference speed, segmentation accuracy, FLOPs, and Params are included.
The FLOPs and Params indicate the number of floating-point operations and the parameters of the network, respectively. Note that our method is also compared with the accuracy-oriented {DeepLab and PSPNet} methods.

{When using the original image (with the size of $1024 \times 2048$) as the input,} our proposed DMA-Net achieves 77.0\% mIoU at the inference speed of 46.7 FPS. Moreover, DMA-Net has only 94.2G FLOPs and 14.60M Params, which are substantially better than some real-time semantic segmentation methods (including SegNet and SQNet). More specifically, DMA-Net is about 32 FPS faster and 20.9\% mIoU higher than SegNet. Although ESPNet achieves the fastest inference speed and the lowest memory consumption, its mIoU is about 16.7\% lower than DMA-Net. Compared with BiSeNet2, DMA-Net not only performs better in terms of accuracy and speed, but also has fewer Params. Although DMA-Net obtains slower inference speed than DFANet, it improves the segmentation accuracy by about 5.7\% mIoU while maintaining the real-time performance. Compared with our previous method
RTHP, DMA-Net adopts higher resolution images as the inputs, and achieves better accuracy (about 3.4\% mIoU higher) and similar inference speed. Furthermore, DMA-Net even achieves better performance than an accuracy-oriented semantic segmentation method. For example, the proposed DMA-Net is about 185 times faster, and about 14\% mIoU higher than DeepLab.

{When using a low-resolution image (with the size of $768 \times 1536$) as the input, our method (denoted as DMA-Net (small)) achieves 75.6\% mIoU at the inference speed of 76.8 FPS. {Compared with SwiftNet, our method not only achieves higher mIoU, but also is nearly 2 times faster.} Therefore, our method achieves a good balance between accuracy and inference speed.}



\begin{figure*}[!tp]
\begin{center}
\includegraphics[width=0.8\linewidth]{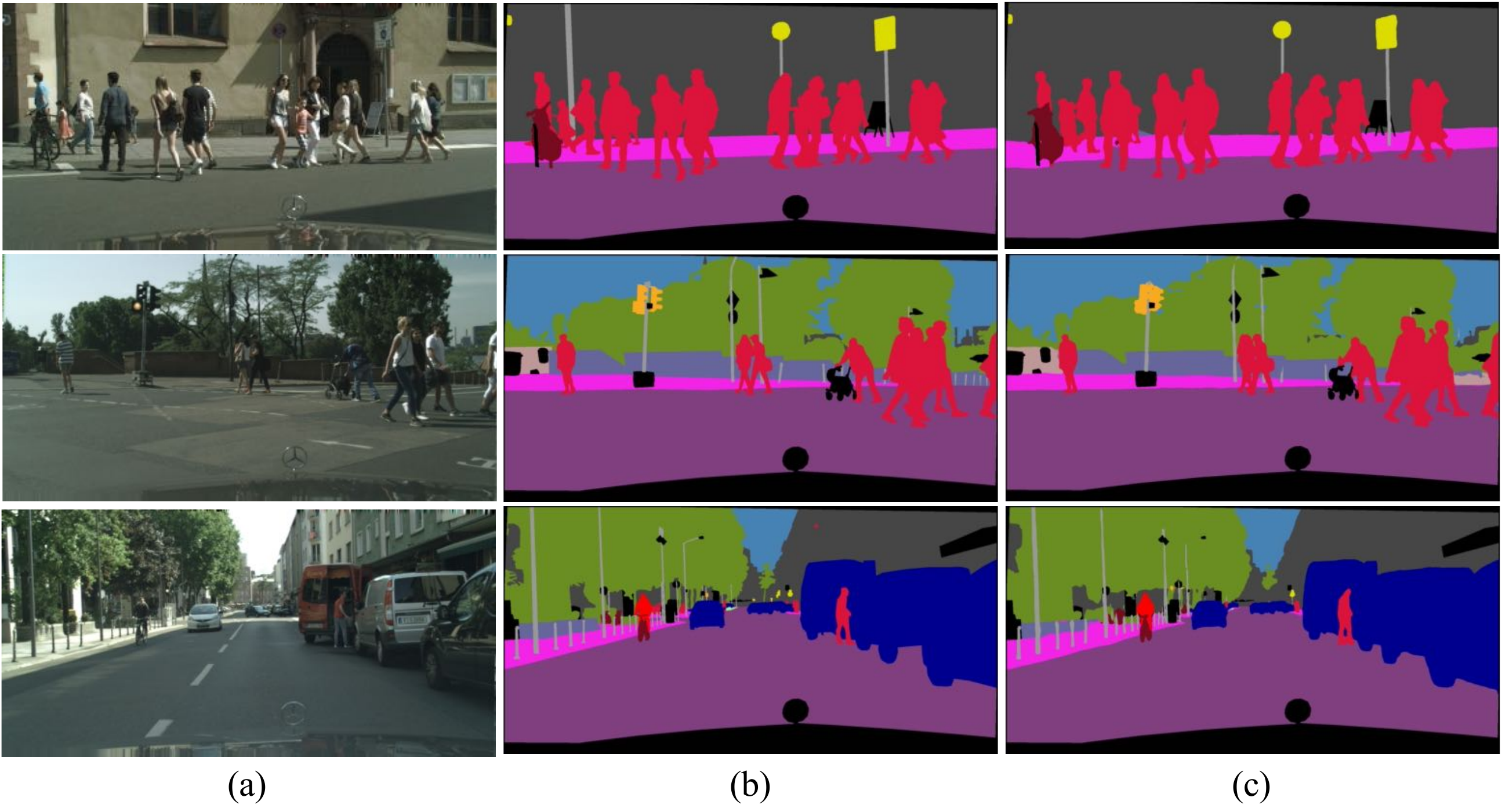}
\end{center}
  \caption{Segmentation results of the proposed DMA-Net on the Cityscapes validation dataset. The images from the first column to the last column respectively denote (a) input images, (b) ground-truth images, and (c) our predicted results.}
\label{fig:Results-cityscapes}
\end{figure*}

\begin{figure*}[!tp]
\begin{center}
\includegraphics[width=0.8\linewidth]{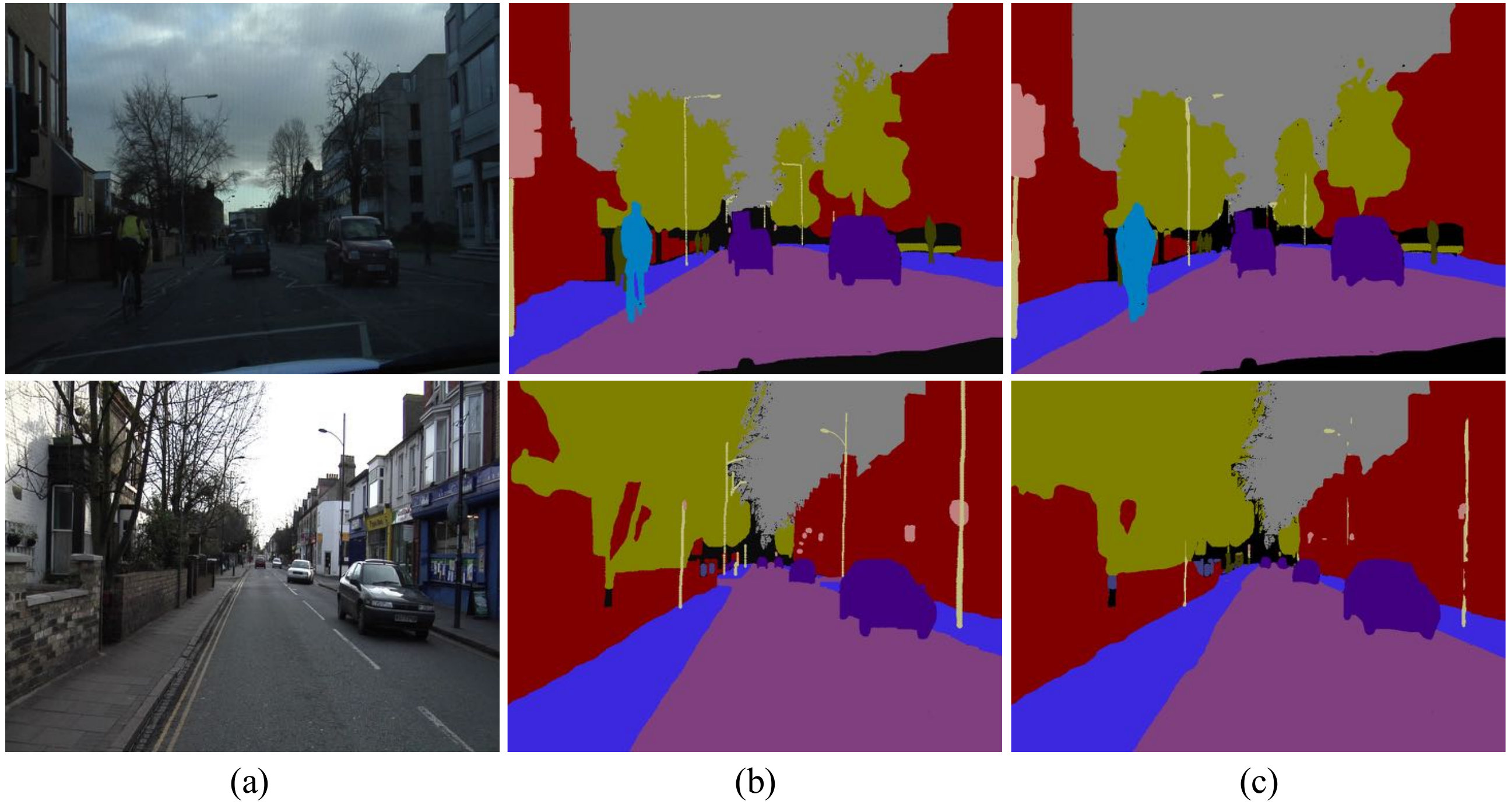}
\end{center}
  \caption{Segmentation results of the proposed DMA-Net on the CamVid test dataset. The images from the first column to the last column respectively denote (a) input images, (b) ground-truth images, and (c) our predicted results.}
\label{fig:Results-camvid}
\end{figure*}

{Similar to BiSeNet, DFANet, and ICNet, DMA-Net also adopts the multi-branch framework. However, compared with BiSeNet that employs a feature fusion module to combine the feature maps from the spatial and context branches, DMA-Net progressively aggregates the feature maps from the high-level branch to the low-level branch. Different from DFANet that performs deep feature aggregation through sub-network and sub-stage cascade, DMA-Net leverages a multi-branch aggregation network (i.e., MAN) based on LERB and FTB. Unlike ICNet that takes the cascade image inputs for different branches, DMA-Net exploits different levels of feature maps from four stages of ResNet-18 as the inputs for multiple branches. Moreover, DMA-Net takes advantage of an elaborately-designed MAN, which not only aggregates different levels of feature maps, but also captures the multi-scale information.}

The per-class, mean-class, and category accuracy values of the Cityscapes test dataset are given in Table \ref{table:compare-two-per-class}. {{Here, the results obtained by {BiSeNet2} are based on the open source codes\footnote{https://github.com/CoinCheung/BiSeNet} and the input image resolution of $1024 \times 2048$.} {It can be seen that our proposed method achieves the best performance on most classes, especially the similar objects {(building vs. wall, truck vs. bus)}.} In particular, our method obtains much higher mIoU than other methods on some classes (such as truck and bus). Although our method obtains the second best performance on some classes (such as vegetation and sky), the difference is trivial (less than 1\% IoU). Meanwhile, our method achieves the lowest mIoU variance, which further shows the effectiveness of our method.}

 {It is worth noting that the Cityscapes dataset was collected from 50 different cities in Germany, where the training set, the validation set, and
the test set consist of the images captured in different cities. Although these subsets show different scene changes, our method is still able to achieve good segmentation performance at real-time inference speed.} Some qualitative segmentation results are shown in Fig. \ref{fig:Results-cityscapes}. Generally speaking, DMA-Net can correctly assign the labels to different scales of objects in street scenes, such as the pedestrians in the second row and the cars in the third row of Fig. \ref{fig:Results-cityscapes}.

{All our experiments are based on an NVIDIA GTX 1080Ti GPU on the desktop platform, which is also employed by state-of-the-art real-time semantic segmentation methods (such as {LEDNet \cite{LEDNet}}, BiSeNet \cite{BiSeNet}, and SwiftNet \cite{SwiftNet}). In this way, we can compare our method with these state-of-the-art methods by using the same platform. As shown in Table \ref{table:compare-two}, our method achieves better segmentation accuracy than other methods at the competitive inference speed. Meanwhile, the number of parameters obtained by our method is only 14.60M. Note that the main differences between the embedded platform in  autonomous driving and the desktop platform are the computing power of graphics cards and memory resources. Therefore, our method is still able to outperform these competing methods when applied to real-world autonomous driving applications.}

\begin{figure*}[!tp]
\begin{center}
\includegraphics[width=0.8\linewidth]{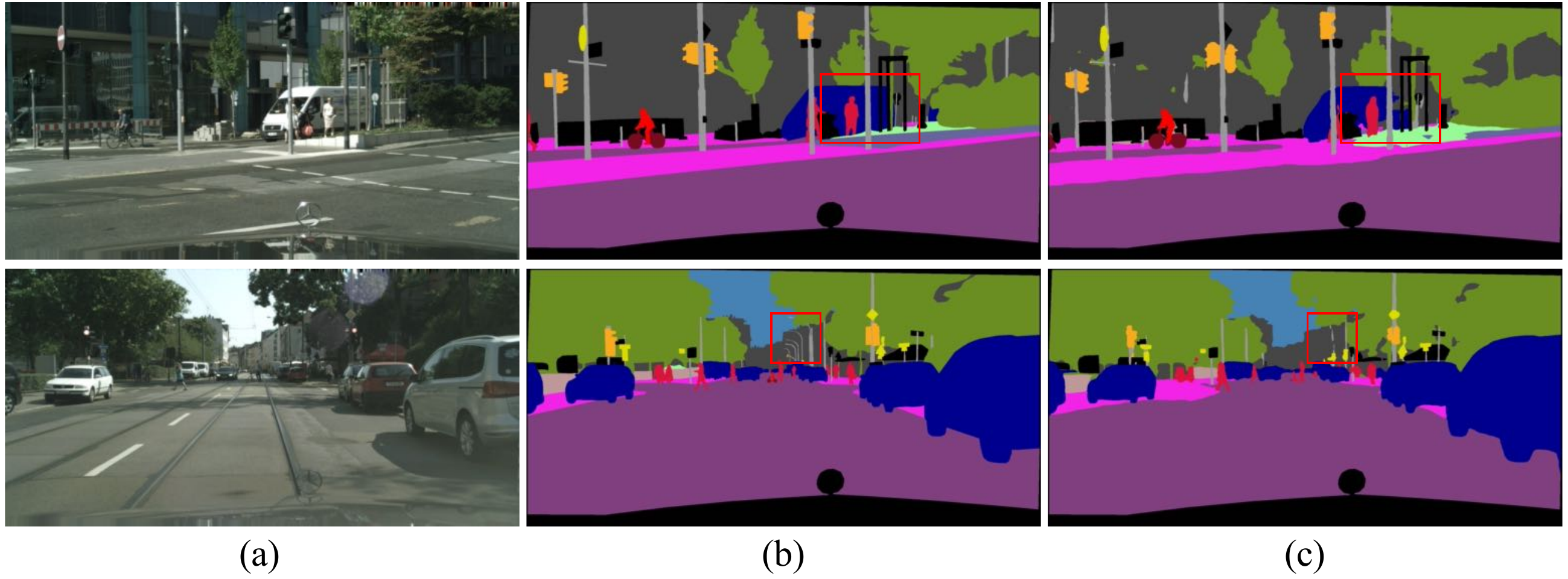}
\end{center}
  \caption{Some failure cases of the proposed DMA-Net on the Cityscapes validation dataset. The images from the first column to the last column respectively denote (a) input images, (b) ground-truth images, and (c) our predicted results.}
\label{fig:failure-Results}
\end{figure*}

\subsection{Results on the CamVid Dataset}
\label{Exp:compare-camvid}
To further illustrate the superiority of our method, we perform experiments on the CamVid dataset.
The evaluation results are reported in Table \ref{Exp:CAMVID}.
{In this experiment, we also fine-tune the model (pre-trained by Cityscapes) on the CamVid dataset to verify the transfer properties of our model. We denote the fine-tuned model as DMA-Net$^\dagger$. }

\begin{table}
    \caption{The Accuracy and Speed Comparison between the Proposed Method and Other Methods on the Camvid Test Dataset. {$\dagger$The Cityscapes Dataset is Used for Pretraining.}}
    \begin{center}
    \setlength{\tabcolsep}{3mm}{
    \renewcommand{\arraystretch}{1}
    \begin{tabular}{lc|cc}
        \whline
        Method & Input Size & mIoU (\%) & Speed (FPS) \\
        \hline\hline
        SegNet \cite{SegNet} & $360\times480$ & 46.4 & 46 \\
        ENet \cite{ENet} & $360\times480$ & 51.3 & 61.2 \\
        ICNet \cite{ICNet} & $720\times960$ & 67.1 & 27.8 \\
        CGNet \cite{cgnet} & $360\times480$ & 65.6 & - \\
        BiSeNet1 \cite{BiSeNet} & $720\times960$ & 65.6 & 175 \\
        BiSeNet2 \cite{BiSeNet} & $720\times960$ & 68.7 & 116.3 \\
        DFANet \cite{DFANet} & $720\times960$ & 64.7 & 120 \\
        SwiftNet \cite{SwiftNet} & $720\times960$ & 72.6 & - \\
        \hline\hline
        DMA-Net (ours) & {$720\times960$} & \textbf{73.6} & \textbf{119.8} \\
        {DMA-Net$^\dagger$ (ours) } & {$720\times960$} & {\textbf{76.2}} & {\textbf{119.8}} \\
        \whline
    \end{tabular}
    }
    \end{center}
    \label{Exp:CAMVID}
\end{table}

{We can observe that our proposed DMA-Net method obtains competitive results (i.e., 73.6\% mIoU at the inference speed of 119.8 FPS) among all the methods. In particular, DMA-Net obtains much faster inference speed than most methods (such as SegNet, ENet, and ICNet).
Compared with BiSeNet2, DMA-Net not only achieves better accuracy (about 4.9\% mIoU higher), but also gives a faster inference speed. Moreover, our DMA-Net also obtains better segmentation performance (about 1\% mIoU higher) than SwiftNet.} In a word, our method achieves a balanced tradeoff between accuracy and speed. {DMA-Net$^\dagger$ achieves the best segmentation accuracy of 76.2\% mIoU, which is about 2.6\% mIoU higher than DMA-Net. This is because the Cityscapes dataset involves a large number of training samples,
enabling us to obtain a powerful pre-trained model. As a result, the pre-trained model can be
easily fine-tuned to classify different classes on the small dataset.}

Note that the images in the CamVid dataset are captured from video sequences. Different from the Cityscapes dataset, there exist {severe illumination variations on CamVid. However, our method still obtains good segmentation results. Some segmentation results are shown in Fig.~\ref{fig:Results-camvid}.} Therefore, our method is robust to scene changes and is applicable to real-world applications requiring real-time inference speed.

\subsection{Limitations}
\label{Limitation}
{In this subsection, we discuss the limitations of our proposed DMA-Net. DMA-Net is able to effectively and efficiently perform semantic segmentation. However, it still suffers from the following two challenges.}

{1)  Severe occlusions between objects. An object can easily be occluded by other objects in street scenes. In particular, when the target object and occluded objects have similar colors and shapes,
our proposed method is prone to give wrong segmentation results.  This is because the similar appearance of different objects makes the network difficult to determine which category the pixel belongs to in the occluded areas.
A failure segmentation result is shown in the first row of Fig.~\ref{fig:failure-Results}. In {the} future, the object depth information can be exploited to address the occlusion problem.}

{2) Small objects in the scenes. Semantic segmentation performs pixel-level classification, where both spatial details and contextual information play an important role in achieving good performance. In our method, the encoder (i.e., ResNet-18) generates different levels of feature maps encoding the spatial and contextual information, while our MAN gradually aggregates the feature maps from the encoder to perform pixel inference. 
In DMA-Net, in order to improve the speed of the network, we do not design a branch to deal with the high-resolution feature maps from  the first sub-network of  ResNet-18.
Therefore, the detailed spatial information of small objects may lose to some extent during feature aggregation. In this way, MAN may not be able to recover the spatial information, thus leading to the misclassification of some small objects in the final segmentation results.  As illustrated in Table \ref{table:compare-two-per-class},  our method achieves a high IoU for some large objects (such as road and building). In contrast, our method gets a low IoU for some small objects (such as fence and pole). A failure segmentation result is shown in the second row of Fig.~\ref{fig:failure-Results}.} In future, more powerful lightweight networks can be designed to provide a good tradeoff between model capacity and inference speed.

{Note that the above two challenges also exist in other real-time semantic segmentation methods (such as ICNet \cite{ICNet} and DFANet \cite{DFANet}).}

\section{Conclusion}
\label{sec:conclusion}

In this paper, we have presented a novel DMA-Net method for real-time semantic segmentation in street scenes. {DMA-Net consists of two main parts: ResNet-18 and MAN. ResNet-18 generates different levels of feature maps, while MAN takes advantage of LERB, FTB, and GCB to aggregate these feature maps and capture the multi-scale information.}
{In particular, LERB makes use of lattice structures to {effectively enhance feature representations} while FTB adaptively generates
the transformed feature maps for feature aggregation.}
Furthermore, GCB encodes the rich global contextual information. These components are tightly coupled and jointly trained to ensure high-quality segmentation results while running at real-time. Experimental results on two challenging street scene benchmarks (including the Cityscapes and the CamVid datasets) have demonstrated the effectiveness and efficiency of our proposed DMA-Net. 

\ifCLASSOPTIONcaptionsoff
  \newpage
\fi




\begin{IEEEbiography}[{\includegraphics[width=1in]{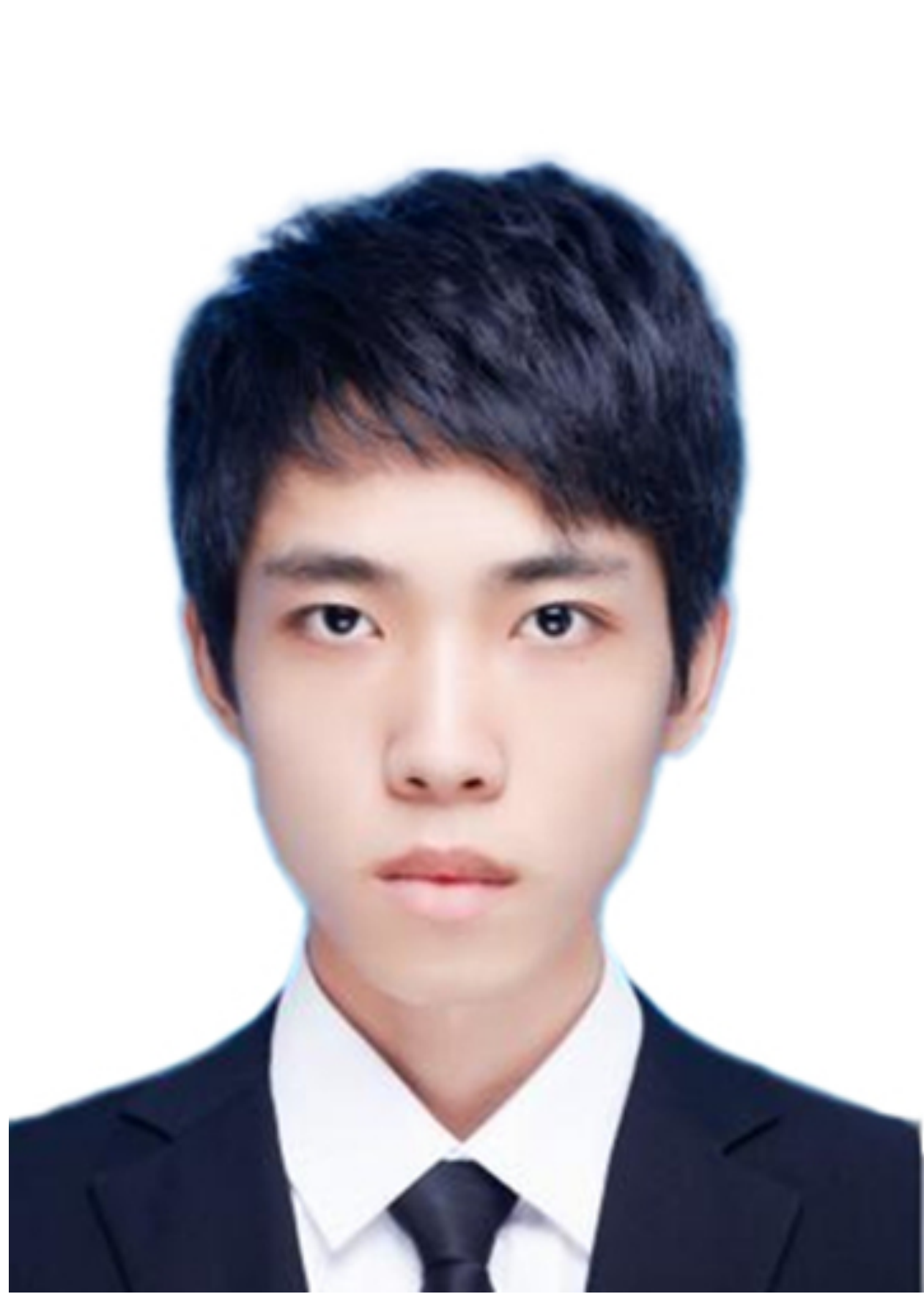}}]{Xi Weng}
is currently a master student in the School of Informatics at Xiamen University, China. His main research interests include deep learning, semantic segmentation, autonomous driving and related fields.
\end{IEEEbiography}

\begin{IEEEbiography}[{\includegraphics[width=1in]{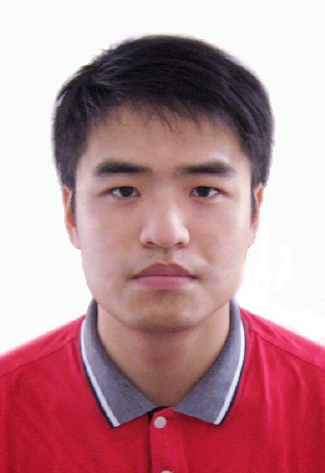}}]{Yan Yan} is currently a professor in the
School of Informatics at Xiamen
University, China. He received the Ph.D. degree
in Information and Communication Engineering from
Tsinghua University, China, in 2009. He worked at Nokia
Japan R\&D center as a research engineer (2009-2010) and Panasonic Singapore Lab as a project leader (2011).
He has published around 100 papers in the international
journals and conferences including the IEEE T-PAMI, IJCV, IEEE T-IP, IEEE T-MM, IEEE T-CYB, IEEE T-CSVT, IEEE T-ITS, IEEE T-AC, PR, CVPR, ICCV, ECCV, ACM MM, AAAI. His research interests include computer vision
and pattern recognition.
\end{IEEEbiography}

\begin{IEEEbiography}[{\includegraphics[width=1in]{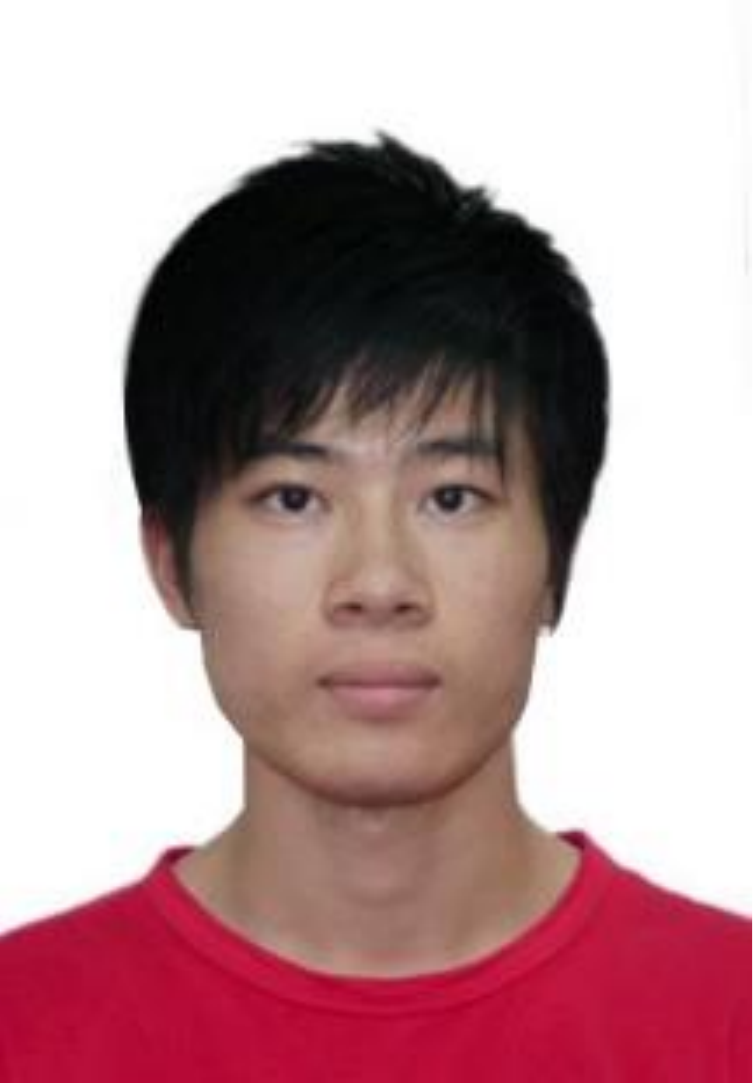}}]{Genshun Dong}
is currently a master student in the School of Informatics at Xiamen University, China. His main research interests include deep learning and semantic segmentation.
\end{IEEEbiography}

\begin{IEEEbiography}[{\includegraphics[width=1in]{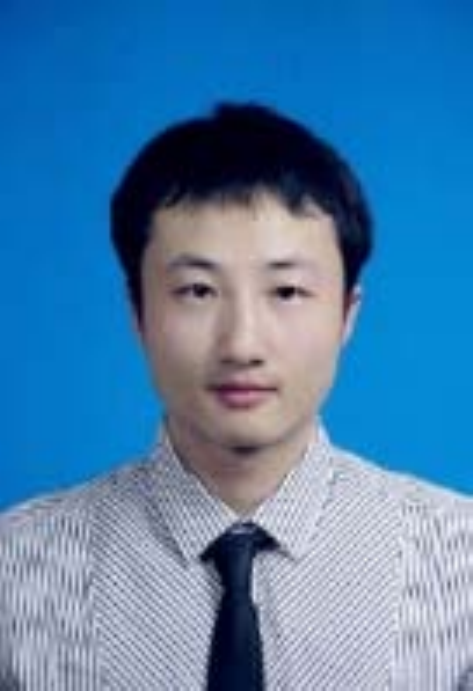}}]{Chang Shu} is currently a lecturer in the School of Information and Communication Engineering at University of Electronic Science and Technology of China, China. He received the Ph.D. degree in Information and Communication Engineering from Tsinghua University, China, in 2011. His research interests include computer vision, machine learning, and pattern recognition.
\end{IEEEbiography}

\begin{IEEEbiography}[{\includegraphics[width=1in]{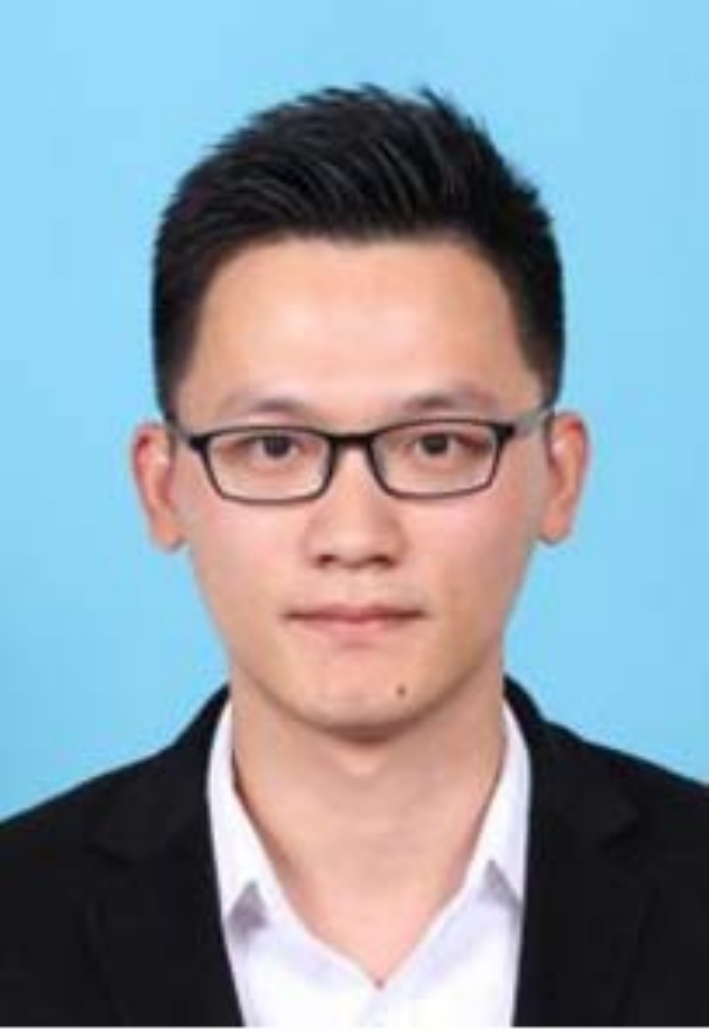}}]{Biao Wang} received his B.E. degree in Electronic Engineering from Dalian University of Technology, in 2012, and the Ph.D. degree in Computer Science from Shanghai Jiao Tong University, in 2020. Currently, he is working as a postdoc in Artificial Intelligence Research Institute in Zhejiang Lab. His research interests include large-scale graph data mining and social network analysis.
\end{IEEEbiography}

\begin{IEEEbiography}[{\includegraphics[width=1in]{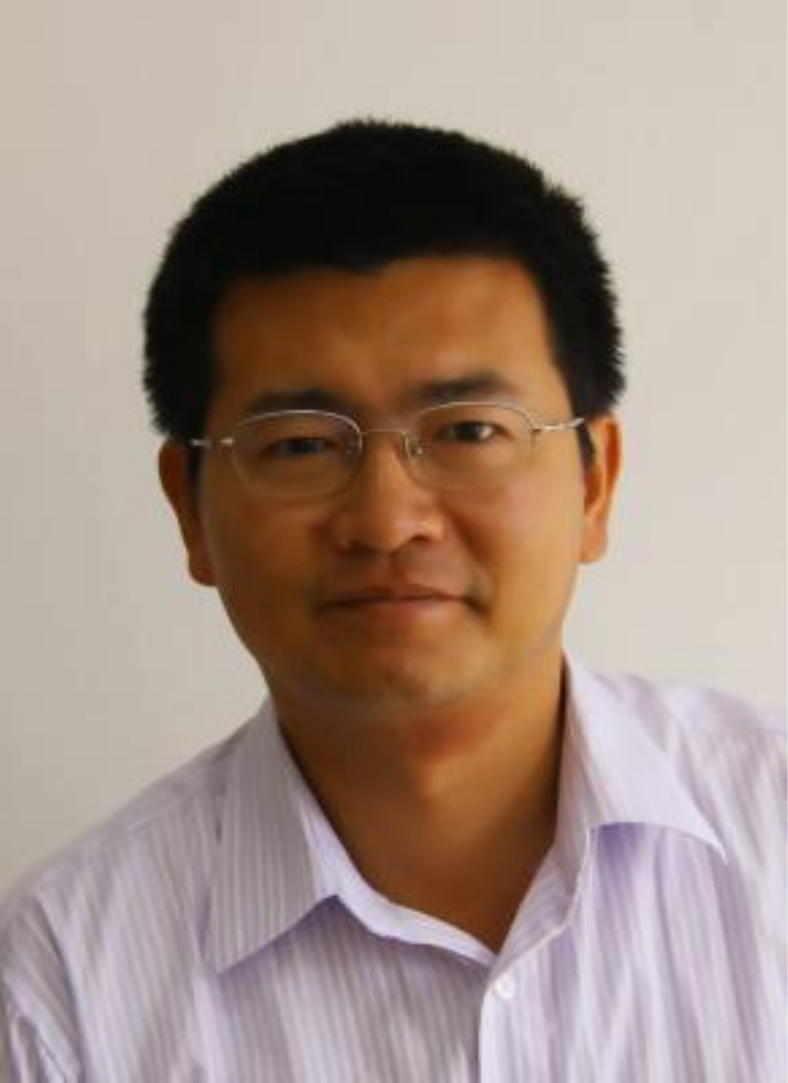}}]{Hanzi Wang}
is currently a Distinguished Professor of
``Minjiang Scholars'' in Fujian province and a Founding
Director of the Center for Pattern Analysis and Machine
Intelligence (CPAMI) at Xiamen University in China.
He received his Ph.D. degree in Computer Vision from
Monash University. His research interests are concentrated
on computer vision and pattern recognition including
visual tracking, robust statistics, object detection,
video segmentation, model fitting, optical flow calculation,
3D structure from motion, image segmentation
and related fields.
\end{IEEEbiography}

\begin{IEEEbiography}[{\includegraphics[width=1in]{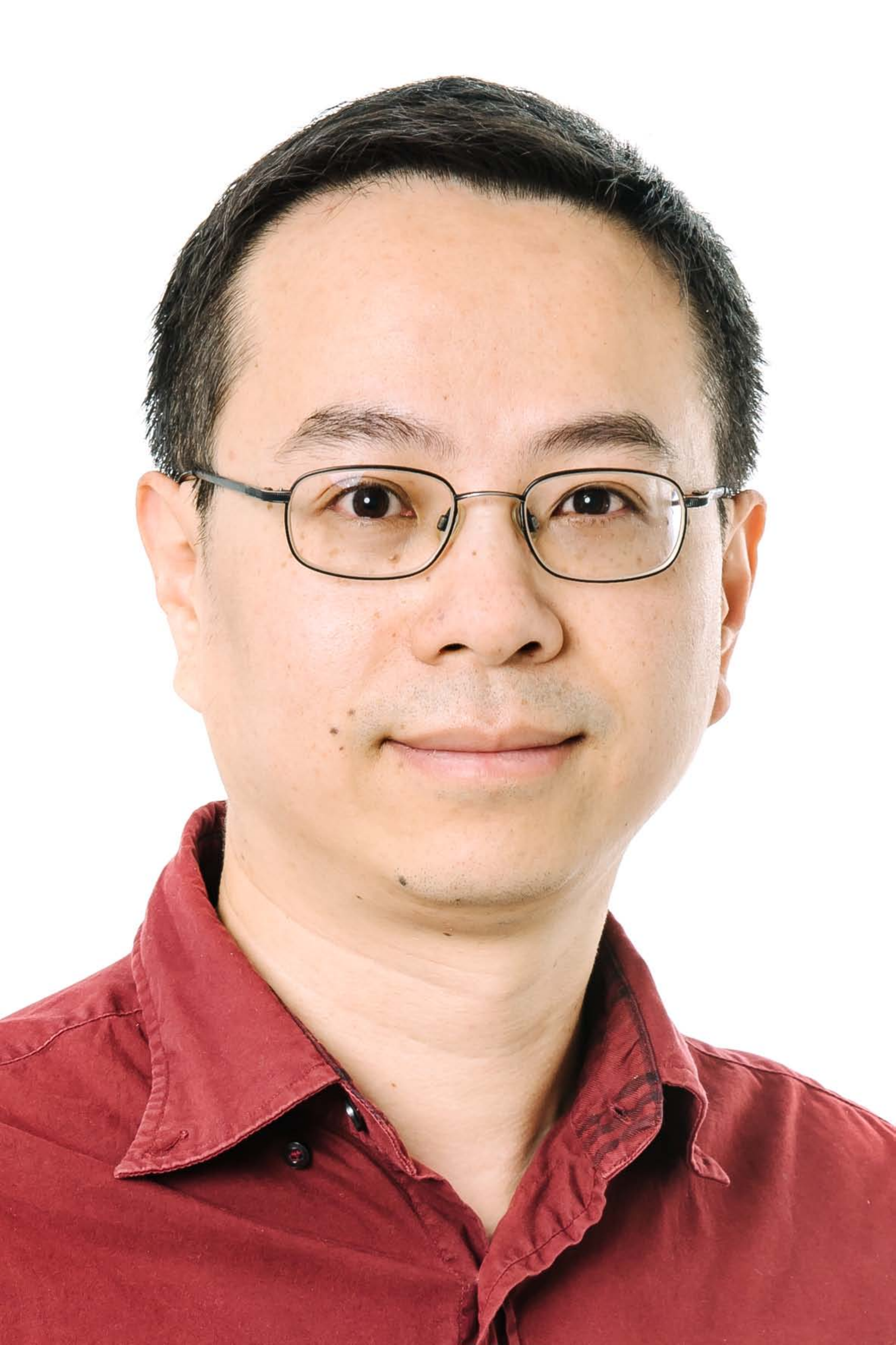}}]{Ji Zhang} is an IET Fellow, IEEE Senior Member, Australian Endeavour Fellow, Queensland International Fellow (Australia) and Izaak Walton Killam Scholar (Canada). He is a Full Professor at USQ and a Visiting Professor at Zhejiang Lab, Tsukuba University, Nanyang Technological University (NTU) and Michigan State University (MSU). Prof. Zhang's research interests include data science, big data analytics, data mining and health informatics. He has published over 230 papers, many appearing in top-tier international journals and conferences, including TKDE, TCYB, TDSC, TKDD, TIST, Information Sciences, KAIS, PRL, WWWJ, JIIS, Bioinformatics, AAAI, IJCAI, SIGKDD, VLDB, ICDE, ICDM, CIKM, WWW Conference, CVPR and COLING. %
 Prof. Zhang is the recipient of Australian Endeavour Award, USQ Research Excellence Award, Head of Department Research Award and three international conference best paper awards.
\end{IEEEbiography}

%


%








\end{document}